\pdfoutput=1

\documentclass[11pt]{article}
\usepackage{CJKutf8} 

\usepackage[final]{acl}

\usepackage{times}
\usepackage{latexsym}
\usepackage{amssymb}
\usepackage{amsfonts}
\usepackage{fdsymbol}
\usepackage{amsmath}
\usepackage[most]{tcolorbox}
\usepackage{afterpage}

\usepackage{url}
\usepackage[T1]{fontenc}

\usepackage[utf8]{inputenc}
\usepackage{multicol}
\usepackage{multirow}
\usepackage{array}
\usepackage{geometry}    
\usepackage{adjustbox}   
\usepackage{tabularx}

\usepackage{microtype}

\usepackage{inconsolata}
\usepackage{soul}

\usepackage{booktabs}
\usepackage{array}
\usepackage{xcolor}
\usepackage{geometry}
\usepackage{tcolorbox}
\usepackage{graphicx}
\usepackage{booktabs}
\usepackage{graphicx}  
\usepackage{xcolor}
\usepackage{subfig}
\usepackage{CJKutf8}

\title{Know You First and Be You Better: \\Modeling Human-Like User Simulators via Implicit Profiles}

\begin{document}
\begin{CJK}{UTF8}{gbsn}
\author{
  \textbf{Kuang Wang\textsuperscript{1}},
  \textbf{Xianfei Li\textsuperscript{1}},
  \textbf{Shenghao Yang\textsuperscript{1}},
  \textbf{Li Zhou\textsuperscript{1}},
\\
  \textbf{Feng Jiang\textsuperscript{2,1}}\thanks{Feng Jiang is the corresponding author and Shenzhen University of Advanced Technology is the corresponding affiliation.},
  \textbf{Haizhou Li\textsuperscript{1,3}}
\\
  \textsuperscript{1}SRIBD, School of Data Science, The Chinese University of Hong Kong, Shenzhen, Guangdong \\
  \textsuperscript{2}Artificial Intelligence Research Institute, Shenzhen University of Advanced Technology \\
  \textsuperscript{3}Department of ECE, National University of Singapore \\
  \texttt{kuangwang@link.cuhk.edu.cn, jiangfeng@suat-sz.edu.cn}
}


\maketitle
\begin{abstract}

User simulators are crucial for replicating human interactions with dialogue systems, supporting both collaborative training and automatic evaluation, especially for large language models (LLMs). However, current role-playing methods face challenges such as a lack of utterance-level authenticity and user-level diversity, often hindered by role confusion and dependence on predefined profiles of well-known figures. In contrast, direct simulation focuses solely on text, neglecting implicit user traits like personality and conversation-level consistency. To address these issues, we introduce the User Simulator with Implicit Profiles (USP), a framework that infers implicit user profiles from human-machine interactions to simulate personalized and realistic dialogues. We first develop an LLM-driven extractor with a comprehensive profile schema, then refine the simulation using conditional supervised fine-tuning and reinforcement learning with cycle consistency, optimizing at both the utterance and conversation levels. Finally, a diverse profile sampler captures the distribution of real-world user profiles. Experimental results show that USP outperforms strong baselines in terms of authenticity and diversity while maintaining comparable consistency. Additionally, using USP to evaluate LLM on dynamic multi-turn aligns well with mainstream benchmarks, demonstrating its effectiveness in real-world applications. We open-source related resources in \url{https://github.com/wangkevin02/USP}.

\end{abstract}

\section{Introduction}

\begin{figure}[t]
   \centering
   \includegraphics[width=0.45\textwidth]{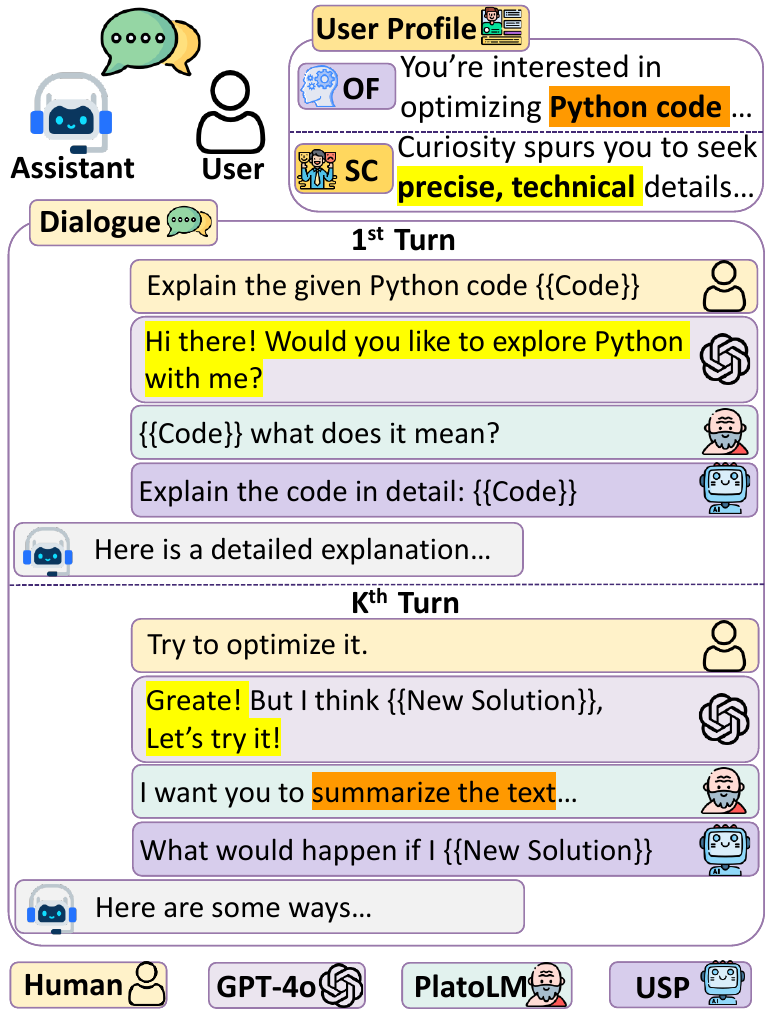}
\caption{Examples of different user simulators in multi-turn human-LLM interactions. "OF" and "SC" represent objective facts and subjective characteristics, respectively. \colorbox{orange}{\rule{0pt}{\fontcharht\font`A}} highlights inconsistencies with the target profile, while \colorbox{yellow}{\rule{0pt}{\fontcharht\font`A}} indicates inauthentic user imitation.}

   \label{fig:intro}
\end{figure}

The user simulator is designed as a proxy for real users in interactions with large language models (LLMs). It can simulate a realistic user by generating the target user's behavior or utterances based on the specified characteristics, enabling dynamic multi-turn interactions with LLMs~\cite{wan2022unified} and scene reproduction~\cite{wang2024depth}. As a result, it becomes an effective alternative~\cite{Liu0YWM0WW23, 0003SM24} in scenarios where real-world human-computer interaction data is difficult to obtain, especially in domains with privacy and ethical concerns, such as medical consultations~\cite{ValizadehP22}. It also helps Simulation-to-Reality (Sim2Real) applications, such as tutorial strategies, election simulations, and public opinion research~\cite{LiuYLC24, zhang2024electionsim, ChuangNSGFYSHR24}.

Recent advances in LLMs have spurred the development of user simulators, improving their naturalness and utility~\cite{0002ZLNC24, ZhangH0LLWLC24}. Mainstream LLM-based role-playing methods~\cite{MoonAKSSBC24} use predefined profiles to mimic diverse user traits. However, as LLMs are typically trained to be universally polite and helpful~\cite{lu-etal-2024-large}, they often lack \textbf{utterance-level authenticity} and struggle with role confusion between user simulation and their inherent assistant nature~\cite{xu2023baize}, as shown in Figure~\ref{fig:intro}. Models like PlatoLM~\cite{KongFWJW24} and Parrot~\cite{SunLZHSZZZG24} address this by training on real human-LLM conversation datasets, but they focus only on text utterances, lacking dialogue control. This limits \textbf{conversation-level consistency} and diverse simulation without seed context. Additionally, they both fail to capture the authentic distribution of \textbf{user-level diversity}, crucial for analyzing group behavior.

To address the issues above, we believe that a user simulator knows users' intrinsic characters hidden in their conversations first, and then can provide a better simulation. Therefore, we treat user simulation as a dialogue reconstruction task and propose a novel framework named the User Simulator with Implicit Profile (USP). It is decomposed into implicit profile extraction to capture the user’s underlying characteristics from the target user dialogue and conditional generation based on the profile.

In this framework, we first propose an LLM-driven profile extractor to extract implicit profiles from user conversations with a well-designed profile schema. Inspired by interpersonal interaction theory\cite{kruglanski2013social}, our profile schema contains two dimensions (objective facts (OF) and subjective characteristics (SC)) with a dozen attributes to describe the user comprehensively. Different from existing works~\cite{abs-2406-13960,tu-etal-2024-charactereval} using attributes as profiles, we further polish the profile attributes into natural, descriptive profiles to ensure generalization.

Then, we integrate the extracted user profiles into the user simulator through two-stage training: (1) conditional supervised fine-tuning with user profiles for utterance-level simulation, and (2) reinforcement learning with cycle consistency to align reflected profiles from simulated dialogues with given profiles for conversation-level simulation. We also implement a diverse profile sampler to capture authentic user distributions.

Our experiments demonstrate that USP improves semantic and stylistic similarity in reconstructed multi-turn dialogues by approximately 34\% and 43\% compared to the leading baseline, with reconstruction errors reduced by half, showcasing enhanced authenticity and diversity. It achieves dialogue profile consistency comparable to GPT-4o (User w/ Profile), improving multi-turn consistency by 14\% while matching single-turn performance. Additionally, USP-based multi-turn dynamic evaluation of LLMs for downstream tasks aligns closely with established benchmarks, enabling finer-grained assessment of LLM performance across diverse user groups. Our key contributions are outlined below:

\begin{itemize} 

\item We propose a novel approach for constructing user simulators using implicit user profiles embedded in human-LLM conversations.

\item We propose a framework that infers implicit user profiles, further enhanced by conditional fine-tuning and reinforcement learning with cycle consistency, to improve simulation at both the utterance and conversation levels.

\item Experiments show that USP outperforms baselines in authenticity and diversity, maintains comparable consistency, and enables effective multi-turn dynamic evaluation of LLMs.

\end{itemize}

\section{Related Works} 

\subsection{General User Simulator}
Early user simulators focused on limited action prediction using agenda-based~\cite{schatzmann2007agenda, schatzmann2009hidden} and model-based methods~\cite{asri2016sequence, kreyssig2018neural}, constrained by early natural language generation capabilities—for instance, generating synthetic binary preferences in conversational recommendation systems~\cite{christakopoulou2016towards}.  

Recent advancements in LLMs enabled more sophisticated simulations of realistic conversations, offering significantly enhanced natural language flexibility. These advances include the use of LLMs for self-chat~\cite{XuGDM23} and dual LLM architectures, where separate models role-play user and assistant based on seed conversations~\cite{DingCXQHL0Z23}. Following these innovations, other trained user simulators, such as PlatoLM~\cite{KongFWJW24} and Parrot~\cite{SunLZHSZZZG24}, learn human discourse patterns directly from human-LLM interactions in conversations. 

\subsection{Persona-based User Simulator}
Since general user simulators often struggle to capture the full spectrum of diverse user needs, it leads a growing interest in persona-based personalization to improve both controllability and diversity in simulations~\cite{takanobu2020multi}. Some researchers attempt to leverage goal generators~\cite{takanobu2020multi} to create diverse user goals or retrieval-based personas derived from historical data~\cite{shi2019build} to guide user simulators in task-oriented dialogue (ToD) systems.

With the rise of LLMs and their strong zero-shot role-playing capabilities~\cite{njifenjou2024role}, prompt-driven user simulation has become the dominant paradigm. LLMs have been used to simulate users with predefined profiles~\cite{ChuangNSGFYSHR24}, model diverse personalities and needs in ToD~\cite{ZhangH0LLWLC24}, and capture user preferences in conversational recommendation~\cite{yoon2024evaluating}. Our method follows this line of work, with a focus on addressing authenticity, consistency, and diversity, which remain underexplored in related studies.

\section{Task Definition} 

We formulate user simulation as a dialogue refactoring task to replicate multi-turn user behavior in a target dialogue \( d_i = \{(u_{i,1}, a_{i,1}), \dots, (u_{i,n}, a_{i,n})\} \), where \( u_{i,j} \) is the \( j \)-th user utterance and \( a_{i,j} \) is the corresponding response answer. Our goal is to achieve high utterance-level and dialogue-level fidelity.  Formally, we minimize utterance-level distance \( D_{\text{utt}}(u_{i,j}, u_{i,j}') \) and dialogue-level distance \( D_{\text{dia}}(d_i, d_i') \), where \( u_{i,j}' \) is the simulated utterance and \( d_i' \) is the simulated dialogue.

Direct simulation struggles to capture personalized traits. Recent studies~\cite{0002ZLNC24, kong2025sharpunlockinginteractivehallucination} demonstrate that role-playing with specific user profiles (\( p_i \)) effectively enables diverse user simulations. However, unlike well-known figures, user profiles in real-world conversations are often implicit and challenging to derive~\cite{WangPQLZWGGN00024}.

To address this, we reformulate the task by extracting the implicit user profile from the dialogue using a profile extractor \( P_{\text{extractor}} \), then reconstructing the target dialogue as Eq.~\ref{eq:task2}.

\begin{equation}\label{eq:task2}
\min_{\substack{d_i' \sim P(\cdot | p_i, \pi_{\theta}) \\ u_{i,j}' \sim P(\cdot | c_{i,j}, p_i, \pi_{\theta})}} \left[ D_{\text{utt}}(u_{i,j}, u_{i,j}') + \alpha D_{\text{dia}}(d_i, d_i') \right],
\end{equation}
where \( p_i = P_{\text{extractor}}(d_i) \), \( \pi_{\theta} \) represents the learnable parameters of the language model, and \( c_{i,j} = \{(u_{i,1}, a_{i,1}), \dots, (u_{i,j-1}, a_{i,j-1})\} \) denotes the ground-truth context up to the \( j \)-th turn. The hyperparameter \( \alpha \) balances the utterance-level and dialogue-level distances.

\section{Modeling User Simulator with Implicit Profiles}
\begin{figure*}[ht]
   \centering
   \includegraphics[width=\textwidth]{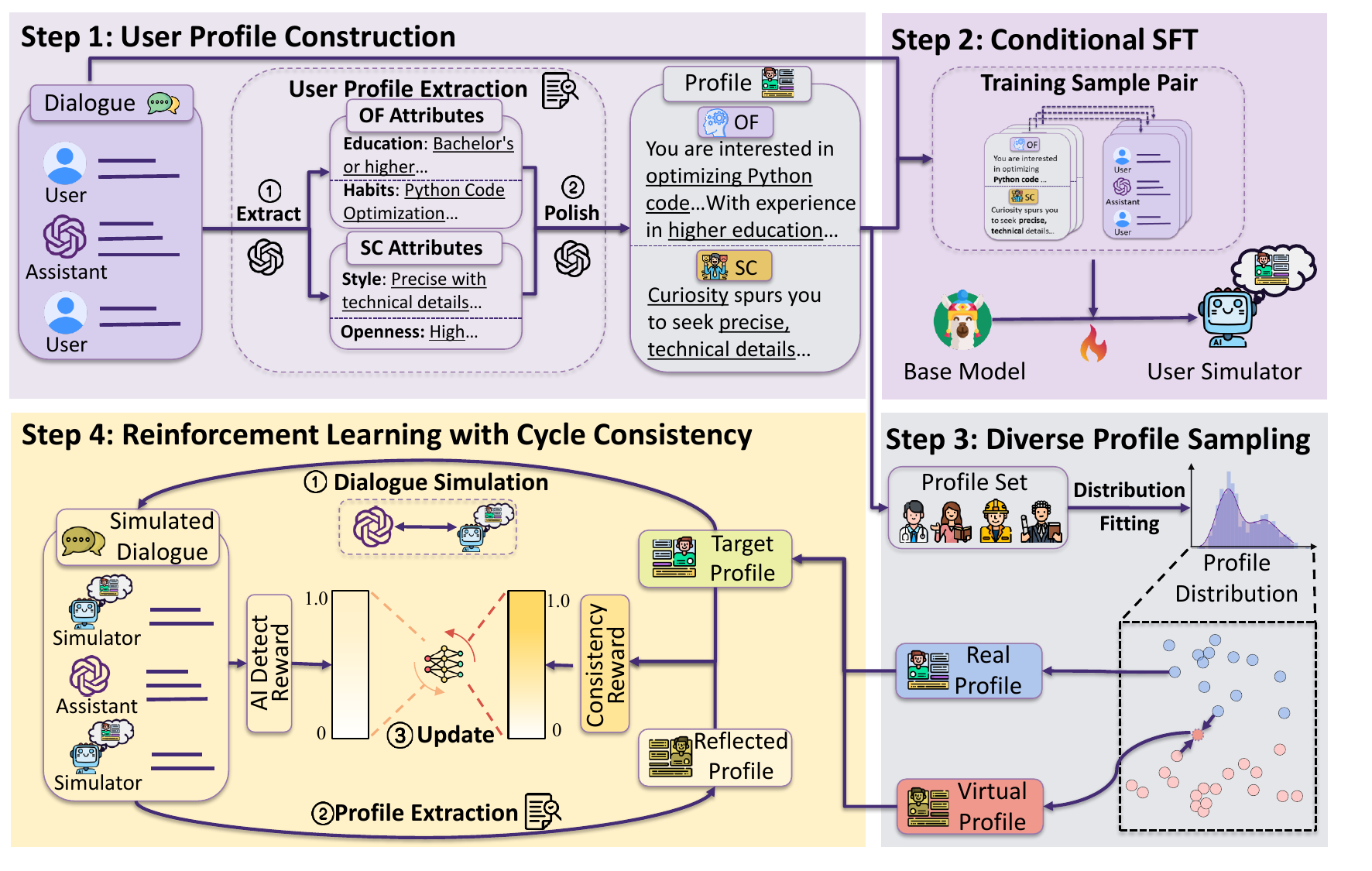}
   \caption{Overview of our proposed  User Simulator with implicit Profile(USP) framework.}
   \label{fig:framework}
\end{figure*}

We propose the User Simulator with Implicit Profiles (USP) framework, shown in Figure~\ref{fig:framework}, to minimize the objective in Eq.~\ref{eq:task2} across four stages. First, we build a user profile extractor with a tailored schema (Section~\ref{sec:profile_construction}). Then, we optimize utterance-level authenticity using conditional SFT (Section~\ref{sec:condition_sft}), ensure conversation-level consistency via Reinforcement Learning with Cycle Consistency (RLCC) (Section~\ref{sec:rlcc}), and achieve user-level diversity through corpus distribution fitting (Section~\ref{sec:profile_sampler}).

\subsection{User Profile Construction}
\label{sec:profile_construction}
\begin{table}[]
\centering
\Huge
\resizebox{\linewidth}{!}{
\begin{tabular}{cll}
\hline
\textbf{Category}                & \multicolumn{1}{c}{\textbf{Dimension}} & \multicolumn{1}{c}{\textbf{Attributes}}   \\ \hline
\multirow{2}{*}{\begin{tabular}[c]{@{}c@{}}Objective \\ Facts \end{tabular}} 
& \begin{tabular}[c]{@{}c@{}}Scene-Consistent \\ Attributes\end{tabular}   & \begin{tabular}[c]{@{}l@{}}Age, Gender, \\ Location, Occupation, \\ Education,  Family Relationship, \\ Routines/Habits, Social \\ Relationships, Other Experiences
\end{tabular} \\ \cline{2-3}
& \begin{tabular}[c]{@{}c@{}}Scene-Related \\ Attributes\end{tabular} & Goals/Plans, Task Details  \\ \hline
\begin{tabular}[c]{@{}c@{}}Subjective \\ Characteristics\end{tabular} & \begin{tabular}[c]{@{}c@{}}Intrinsic \\ Characteristics\end{tabular} & \begin{tabular}[c]{@{}l@{}}Big Five Personality Traits, \\ Language Styles \end{tabular} \\  \hline

\end{tabular}
}
\caption{The Designed User Profile Schema.}
\label{tab:persona}
\end{table}

\subsubsection{User Profile Schema} 
We believe that the user profile should reveal user characteristics from two aspects: explicit personal information and implicit communication styles. Therefore, inspired by interpersonal interaction theory~\cite{kruglanski2013social}, we design a user profile schema containing objective facts(OF) and subjective characteristics(SC) to represent them, as shown in Table~\ref{tab:persona}. 

The OF focuses on common topics in human conversation~\citep{abs-2406-13960,dunbar1997human}, including scene-consistent attributes (such as age, gender, and location) and scene-related attributes (such as the goal and task details). SC encompasses external personality dimensions, reflected in language style~\cite{WangPQLZWGGN00024}, and internal personality dimensions, captured by the Big-Five Traits~\cite{gosling2003very}.

Unlike prior work~\cite{abs-2406-13960, tu-etal-2024-charactereval} that relies on discrete attributes for user profiles, we further reformulate these attributes into coherent narrative descriptions to enhance generalization and flexibility.

\subsubsection{User Profile Extractor} 
\label{sec:profile_construction_flow}

To obtain such a user profile, we design an LLM-driven user profile extractor that extracts the implicit user profile from the human-LLM conversation. The extractor first leverages advanced LLM (such as GPT-4o) to extract the user character attributes mentioned above with a well-designed prompt. Then, the extractor collects the valid attributes together and polishes them into natural language descriptions. Further prompt details regarding the extractor can be found in Appendix~\ref{appendix:profile_dataset}.

\subsubsection{Profile Quality Verification}

Existing role-playing methods rely on predefined profiles and dialogues from separate sources, either extracted from novel segments or generated by LLMs~\cite{zhou2024characterglm, chen2023large, characterllm} without verifying alignment~\cite{WangPQLZWGGN00024, abs-2406-13960}. This overlooks the correlation between profiles and dialogues, potentially hindering simulation performance~\cite{yu2024beyond}. To provide an automatic metric for evaluating this, we propose Dialogue Profile Consistency (DPC), which frames consistency as a retrieval task~\cite{JandaghiSBPS24}. DPC employs an F1-score approach, assessing consistency through atomic fact verification by measuring both precision (\(\text{DP.P}\)) and recall (\(\text{DP.R}\)).

We first introduce Factual Consistency (Fact.Con), an adaptation of FactScore~\cite{MinKLLYKIZH23} tailored for dialogue scenarios, as defined in Eq.~\ref{eq:fact_con}. Given a target \(T\), we assess its consistency with the source by decomposing \(T\) into atomic facts \(af_k\) using an atomic fact generator (\(\text{afg}\)). We then compute the natural language inference (NLI) score for each atomic fact with respect to the source \(S\).

\begin{equation}\label{eq:fact_con}
\text{Fact.Con}(S, T) = \frac{1}{|af_k|} \sum_{af_k \in \text{afg}(T)} \text{NLI}(S, af_k)
\end{equation}

where \(\text{NLI}(\cdot,\cdot)\) denotes the NLI model, implemented using prompt-based GPT-4o.

We then define \(\text{DP.P}_i = \text{Fact.Con}(d_i, p_i)\), which measure the accuracy of profile description information and \(\text{DP.R}_i = \text{Fact.Con}(p_i, d_i)\) to assess the profile’s coverage of the dialogue. The DPC is their harmonic mean. When dialogue \(d_i\) serves as the target \(T\), each user utterance \(u_{i,j}\) is treated directly as an atomic fact \(af_k\). Conversely, when the profile serves as the target \(T\), we utilize \(\text{afg}\) followed~\cite{MinKLLYKIZH23} to decompose it into atomic facts. 

Additionally, we use a Validation Score (Val.Score) to evaluate SC description quality based on the dialogue, rated on a 1–5 scale using GPT-4o(prompts detailed in Appendix~\ref{appendix:prompt_design}).

\subsection{Conditional Supervised Fine-Tuning}
\label{sec:condition_sft}

To empower the LLM with the general capability to simulate diverse users at the utterance level, we utilize conditional supervised fine-tuning based on user profiles. It enables the LLM to learn the conditional generation mapping based on both the extracted profile \(p_i\) and context \(c_{i,j}\). As a subtle misalignment between the core objectives of the user simulator and the response model, the SFT language modeling loss focuses on optimizing user utterance as shown in Eq.~\ref{eq:lm_loss}.

\begin{equation}\label{eq:lm_loss}
\min_{\pi_\theta} \sum_{i,j,k} -\log P(u_{i,j,k} | u_{i,j,<k}, c_{i,j}, p_i, \pi_\theta)
\end{equation}
where $u_{i,j,k}$ denotes the $k$-th token of the $u_{i,j}$.


\subsection{Diverse Profile Sampling}
\label{sec:profile_sampler}

We propose Diverse Profile Sampling to generate naturalistic user profiles that reflect real-world characteristic distributions. Our method first embeds constructed profiles into a semantic space using SimCSE~\cite{simcse}, followed by dimensionality reduction via UMAP~\cite{mcinnes2018umap}. We then apply Gaussian Kernel Density Estimation (GKDE) to fit the underlying distribution, allowing probabilistic sampling of realistic profiles for downstream tasks such as majority representation. To further enhance diversity, we synthesize virtual profiles by combining OF and SC descriptions from nearest neighbors, producing novel yet plausible profile variants.

\subsection{Reinforcement Learning with Cycle Consistency}
\label{sec:rlcc}
The conditional SFT stage enables the user simulator to generate human-like utterances focused on forward consistency, producing precise responses aligned with the target profile. However, it does not ensure full reflection of the profile, i.e., backward consistency or profile recall. To overcome this and improve conversation-level consistency, we introduce Reinforcement Learning with Cycle Consistency (RLCC), which enhances alignment between the user simulator’s actual behavior reflected in simulated dialogues and the target behavior defined by the profile, ensuring a closer match to the intended target profile.

In this stage, the user simulator $u_i'$ interacts with a response LLM based on a target profile $p_i$, sampled via Diverse Profile Sampling, to generate a simulated dialogue $d_i'$. The dialogue ends when it reaches the maximum context length or a predefined turn limit (set to 10). The reflected profile $p_i'$ is then extracted from $d_i'$ using the profile generator. Our goal is to maximize the semantic similarity between the target profile $p_i$ and the reflected profile $p_i'$, both in objective facts and subjective characteristics, as defined in Eq.~\ref{eq:reward_cc}. 
\begin{align}\label{eq:reward_cc}
\max_{\pi_{\theta}} \; \mathbb{E}_{p_i \sim D, d_i' \sim \pi_\theta(p_i)}\left[ \text{Sim}(p_i, P_{extractor}(d_i')) \right]
\end{align}

where $\text{Sim}(\cdot,\cdot)$ is a similarity model (SimCSE~\cite{simcse}), and $D$ denotes the virtual profiles dataset from the sampler in Section~\ref{sec:profile_sampler}. The dialogue-level reward is uniformly attributed to each user utterance, defined as $r_{i,j}^{cc} = \text{Sim}(p_i, P_{extractor}(d_i'))$.

To prevent reward hacking, the AI detection reward is included as an auxiliary component. The final reward, defined in Eq.~\ref{eq:reward_define}, is utilized to optimize profile recall via Proximal Policy Optimization (PPO)\cite{schulman2017proximal}.

\begin{equation}
r_{i,j} = \lambda r^{cc}_{i,j} + (1-\lambda) r^{ai\_detect}_{i,j}
\label{eq:reward_define}
\end{equation}

where \( r^{ai\_detect}_{i,j} = \text{AI\_detect}(u_{i,j}') \), and \( \lambda = 0.8 \) prioritizes cycle consistency. The \(\text{AI\_detect}\) refers to a binary AI detection model~\cite{yang2024chatgpt} that predicts the probability of an utterance being AI-generated. Both the AI detection model and profile generator are fine-tuned on our training dataset, with details provided in Appendix~\ref{appendix:model_setup}.

\section{Experiments}
We evaluate user simulators on authenticity, consistency, and multi-turn continuity at both utterance and conversation levels, while measuring diversity by comparing the dialogue distributions of simulated and real users.

\subsection{Datasets}

We select the popular LMSYS-Chat-1M~\cite{zheng2023lmsys}, which contains one million human-LLM conversations. Following prior work~\cite{KongFWJW24}, we filter out non-English, toxic, and redundant samples, resulting in 94,874 conversations (87,882 for training, 4,626 for validation, and 2,366 for testing). Each conversation is then annotated with user profiles using the GPT-4o-based extractor described in Section~\ref{sec:profile_construction}, forming the LMSYS-USP dataset. Detailed preprocessing steps are provided in Appendix~\ref{appendix:preprocess}.

\begin{table*}[!ht]
    \centering
    \small
    \begin{tabular}{llcccccc}
    \toprule
    \multirow{2}{*}{\textbf{Dataset}} & \multirow{2}{*}{\textbf{Profile Source}} & \multicolumn{5}{c}{\textbf{OF}} & \multicolumn{1}{c}{\textbf{SC}} \\
    \cmidrule(lr){3-7} \cmidrule(lr){8-8}
    & & DP.P$\uparrow$  & Avg DP.P \# Fact  & DP.R$\uparrow$  & Avg DP.R \# Fact  & DPC$\uparrow$  & Val.Score$\uparrow$  \\
    \midrule
    LMSYS-USP& GPT4o & 86.89 & 25.64 & 82.24 & 3.71 & 84.50 & 4.42 \\
    LMSYS-USP & Distill-llama3 & 86.15 & 23.81 & 81.95 & 3.71 & 84.00 & 4.36 \\
    Persona Chat & GPT4o & 86.21 & 22.82 & 62.76 & 7.86 & 72.64 & 4.35 \\
    Persona Chat & Human & 76.21 & 8.59 & 42.94 & 7.86 & 54.93 & - \\
    ConvAI2 & GPT4o & 68.71 & 17.44 & 39.15 & 9.97 & 49.88 & 3.47 \\
    ConvAI2 & Human & 25.69 & 8.70 & 12.64 & 9.97 & 16.94 & - \\
    \bottomrule
    \end{tabular}
    \caption{Automated evaluation of profile quality across datasets. Avg DP.P \# Fact denotes the average number of atomic facts per user profile, while Avg DP.R \# Fact  represents the average number of user utterances per dialogue. Note that human-annotated profiles in PersonaChat and ConvAI2 contain no subjective characteristics.}
    \label{tab:profile_auto_eval}
\end{table*}

We use DPC and Val.Score to automatically evaluate the quality of extracted user profiles on 100 randomly selected samples from the LMSYS-USP test set, as well as on 100 samples each from Persona-Chat~\cite{KielaWZDUS18} and ConvAI2\footnote{We use the human-to-bot dataset from \url{https://huggingface.co/datasets/convai-challenge/conv_ai_2}}~\cite{DBLP:journals/corr/abs-1902-00098}, which include manually annotated predefined profiles. As shown in Table~\ref{tab:profile_auto_eval}, the extracted profiles achieve over 84\% DPC, with even distill-llama3 results comparable to GPT-4o, demonstrating the effectiveness of our annotation method. Manual evaluation further confirms profile quality, with average scores exceeding 4 out of 5 (see Appendix~\ref{appendix:huamn_eval} for details).

\subsection{Configurations}
We train USP based on  LLaMA-3-8B-Base~\cite{llama3modelcard} model. The conditional SFT is conducted on the training dataset using 4 A100 40GB GPUs, with full fine-tuning over 3 epochs at a learning rate of 5e-5 and max length set to 4096, taking approximately two days. Our diverse profile sampler then randomly selects 1,000 samples from the training set for virtual user sampling, combining objective facts and subjective characteristics to generate about 1 million profiles. From these, we select the 5,000 profiles least similar to the training dataset for the RLCC phase. RLCC training uses two H20 96GB GPUs for 5 days, utilizing a KL coefficient of 0.01, a learning rate of 5e-7, and training for 1 epoch.

\subsection{Baseline Models} 

\textbf{(1) User Simulator without User Profile}: This includes untrained GPT-4o (User w/o Profile), which use GPT-4o to predict user utterances based solely on context, and PlatoLM~\cite{KongFWJW24}, a baseline fully fine-tuned on our dataset using LLaMA-3-8B-Base, representing a profile-agnostic approach.

\textbf{(2) User Simulator with User Profile}: We employ  GPT-4o (User w/ Profile) and LLaMA3 (User w/ Profile)  leveraging GPT-4o and LLaMA-3-8B-Instruct~\cite{llama3modelcard} as profile-conditioned role-playing agents, alongside CharacterGLM~\cite{zhou2024characterglm}, a flexible profile-based baseline, and CharacterLLM~\cite{characterllm}, designed to emulate public figures.

\subsection{Metrics}
\textbf{Authenticity:} We evaluate semantic and stylistic similarity using SimCSE~\cite{simcse} and style embeddings~\cite{WegmannSN22}, respectively, to compute $D_{\text{utt}}(u_{i,j}, u'_{i,j})$ and $D_{\text{dia}}(d_i, d_i')$. To assess stylistic consistency, we report Author Verification Accuracy (AVA)~\cite{WegmannSN22}, which measures whether sentence pairs are attributed to the same author based on similarity thresholds. Dialogue-level distances are computed by concatenating all user utterances.

\textbf{Consistency:} We evaluate profile-based generation consistency using reverse metrics: r-DP.P and r-DP.R. Unlike the DPC series, which treats dialogue as ground truth to assess profile quality, these metrics measure factual alignment from the profile’s perspective. Specifically, r-DP.P is defined as $\text{Fact.Con}(p_i, d_i')$, and r-DP.R as $\text{Fact.Con}(d_i', p_i)$. Their harmonic mean, r-DPC, captures overall consistency. For utterance-level analysis, we report the average DP.P. Additionally, we use Persona Coverage (P.Cover)~\cite{SongZCWL19} for keyword match and the GPT-4o-rated Subjective Characteristic Score (SC.Score) (see Appendix~\ref{appendix:prompt_design}) to assess subjective trait expression performance.

\textbf{Diversity:}
We measure the Absolute Difference Value (ADV), defined as the Euclidean distance between PCA-reduced embeddings of generated and target dialogues, to quantify the distributional discrepancy between simulated and target dialogues.

\textbf{Continuity:} Multi-turn dialogue continuity ability is evaluated via the early stop rate (ESR), which detects premature endings triggered by repetitive responses or repeated expressions of gratitude across three turns.

\subsection{Results}

\subsubsection{Utterance-Level Evaluation}

\begin{table*}[!ht]
    \centering
    \setlength{\tabcolsep}{6pt}
    \tiny
    \resizebox{\linewidth}{!}{
    \begin{tabular}{l l *{6}{c}}  
    \toprule
    \multirow{2}{*}{\textbf{Model Type}} & \multirow{2}{*}{\textbf{Model}} & 
    \multicolumn{3}{c}{\textbf{Authenticity}} & \multicolumn{3}{c}{\textbf{Consistency}} \\ 
    \cmidrule(lr){3-5} \cmidrule(lr){6-8}
    & & Sem-Sim$\uparrow$ & Style-Sim$\uparrow$ & AVA$\uparrow$ & r-DP.P$\uparrow$ & P.Cover$\uparrow$ & SC.Score$\uparrow$ \\ 
    \midrule
    \multirow{2}{*}{w/o Profile} 
    & GPT-4o (User w/o Profile) & 40.24 & 13.75 & 11.28 & -- & -- & -- \\ 
    & PlatoLM & 39.37 & 43.11 & 40.29 & -- & -- & -- \\ 
    \midrule
    \multirow{5}{*}{w/ Profile} 
    & Character\_LLM & 37.54 & 18.88 & 15.03 & 54.77 & 66.62 & 2.43 \\ 
    & Character\_GLM & 38.51 & 22.28 & 18.17 & 68.72 & 57.72 & 2.95 \\ 
    & LLaMA3 (User w/ Profile) & 39.82 & 14.88 & 13.47 & 82.19 & 72.29 & 3.92 \\ 
    & GPT-4o (User w/ Profile) & 41.66 & 5.74 & 9.87 & \textbf{92.73} & \textbf{73.34} & \textbf{4.71} \\ 
    & USP (w/o RLCC) & \textbf{54.25} & 46.57 & \textbf{43.61} & 71.30 & 71.56 & 3.36 \\ 
    & USP & 53.38 & \textbf{46.60} & 43.35 & 72.61 & 71.23 & 3.39 \\ 
    \bottomrule
    \end{tabular}
    }
    \caption{Utterance-level performance comparison of different user simulator.}
    \label{tab:one_turn}
\end{table*}

In utterance-level evaluation, we assess the quality of single-turn responses generated by different user simulators given the golden context.

As shown in Table~\ref{tab:one_turn}, USP outperforms all baselines in authenticity, as measured by both semantic similarity (Sem-Sim: 53.38) and stylistic similarity (Style-Sim: 46.60). This shows the effectiveness of our implicit profile-based approach for user-LLM dialogue reconstruction, especially compared to non-profile baselines like PlatoLM. While dedicated role-playing models (e.g., GPT-4o (User w/ Profile)) achieve higher consistency scores (r-DP.P) due to direct profile keyword copying with high P.Cover (73.34), our USP strikes a better balance between authenticity and consistency, as shown by the intuitive examples in Section~\ref{sec:case_study}.

\subsubsection{Conversation-Level Evaluation}
\begin{table*}[!ht]
    \centering
    \resizebox{\linewidth}{!}{
    \begin{tabular}{l l *{10}{c}}
    \toprule
    \multirow{2}{*}{\textbf{Model Type}} & \multirow{2}{*}{\textbf{Model}} & \multicolumn{1}{c}{\textbf{Continuity}} & \multicolumn{3}{c}{\textbf{Authenticity}} & \multicolumn{5}{c}{\textbf{Consistency}} \\ 
    \cmidrule(lr){3-3} \cmidrule(lr){4-6} \cmidrule(lr){7-11}
    & & ESR$\downarrow$ & Sem-Sim$\uparrow$ & Style-Sim$\uparrow$ & AVA$\uparrow$ & r-DP.P$\uparrow$ & r-DP.R$\uparrow$ & r-DPC$\uparrow$ & P.Cover$\uparrow$ & SC.Score$\uparrow$ \\ 
    \midrule
    \multirow{2}{*}{w/o Profile} 
    & GPT-4o (User w/o Profile) & 35 & 48.91 & 14.21 & 10.58 & -- & -- & -- & -- & -- \\ 
    & PlatoLM & 18 & 43.24 & 32.43 & 31.60 & -- & -- & -- & -- & -- \\ 
    \midrule
    \multirow{5}{*}{w/ Profile} 
    & Character\_LLM & 52 & 23.37 & 7.13 & 4.69 & 25.48 & 6.43 & 10.27 & 21.49 & 2.82 \\ 
    & Character\_GLM & 44 & 40.19 & 10.86 & 12.67 & 39.51 & 29.61 & 33.85 & 42.75 & 3.64 \\ 
    & LLaMA3 (User w/ Profile) & 31 & 46.84 & 10.58 & 11.63 & 67.09 & 29.98 & 41.44 & 47.72 & 4.19 \\
    & GPT-4o (User w/ Profile) & 32 & 48.87 & 10.15 & 11.26 & \textbf{76.59} & 43.72 & 55.66 & \textbf{51.02} & \textbf{4.56} \\ 
    & USP (w/o RLCC) & 12 & \textbf{66.17} & 40.01 & 35.68 & 53.17 & 71.88 & 61.13 & 42.63 & 3.24 \\ 
    & USP & \textbf{10} & 65.39 & \textbf{46.23} & \textbf{38.77} & 56.24 & \textbf{74.38} & \textbf{64.05} & 44.08 & 3.35 \\ 
    \bottomrule
    \end{tabular}
    }
    \caption{Conversation-level performance comparison of different user simulators.}
    \label{tab:multi_turn}
\end{table*}

In the conversation-level evaluation, we assess the quality of multi-turn dialogues generated by different user simulators interacting with GPT-4o, each provided with either a given profile or the first turn of a reference dialogue.

As shown in Table~\ref{tab:multi_turn}, USP outperforms baseline models in authenticity, consistency, and continuity. With the lowest ESR (10), USP ensures superior dialogue continuity, avoiding issues like repetitive generation and reciprocal appreciation loops seen in baselines. Its advantage in authenticity is especially evident in multi-turn scenarios, compared to sentence-level evaluations. In terms of consistency, USP excels with a high r-DP.R (74.38) and significantly better r-DPC (64.05), demonstrating strong conditional generation consistency. Unlike role-playing models such as GPT-4o (User w/ Profile) and LLaMA3 (User w/ Profile), which show high P.Cover and r-DP.P but lower overall profile dialogue consistency, USP demonstrates a deeper and more comprehensive understanding of user behavior, moving beyond surface-level keyword matching to deliver a more vivid user simulation.

\subsubsection{Human Evaluation}

We randomly selected 100 samples from the test set and engaged 8 evaluators to assess conversations on authenticity and consistency. Authenticity was evaluated based on Style, Semantics, and Quality, while consistency focused on Accuracy, Completeness, and Quality. Detailed criteria are in Appendix~\ref{appendix:huamn_eval}. 

Table~\ref{tab:human-eval} shows USP’s clear superiority in both authenticity and consistency. USP outperforms GPT-4o (User w/ Profile) in authenticity (74 vs. 13) and consistency (61 vs. 35). It also surpasses PlatoLM trained on the same data in authenticity, demonstrating the advantage of implicit profile modeling. The larger gap in consistency (43 vs. 30) compared to authenticity (37 vs. 31) between USP and USP (w/o RLCC) highlights RLCC’s key role in aligning profiles with dialogues.

\begin{table}[!ht]
    \centering
    \small
    \begin{tabular}{lcc}
    \toprule
    \multirow{2}{*}{Baseline} & \multicolumn{2}{c}{Metrics (\% USP win/tie/loss)} \\
    \cmidrule(lr){2-3}
    & Authenticity & Consistency \\
    & ($\kappa$=0.548) & ($\kappa$=0.561) \\
    \midrule
    GPT-4o (User w/ Profile) & 74/13/13 & 61/4/35 \\
    PlatoLM & 55/12/33 & - \\
    USP (w/o RLCC) & 37/32/31 & 43/27/30 \\
    \bottomrule
    \end{tabular}
    \caption{Human evaluation of USP win rates over baselines in terms of authenticity and consistency. $\kappa$ denotes the within-group kappa coefficient. Note that PlatoLM, as a non-profile baseline, contain no consistency results.}

    \label{tab:human-eval}
\end{table}

\subsubsection{Diversity Sampling Evaluation}

\begin{figure}[!ht]
\centering
\includegraphics[width=0.95\linewidth]{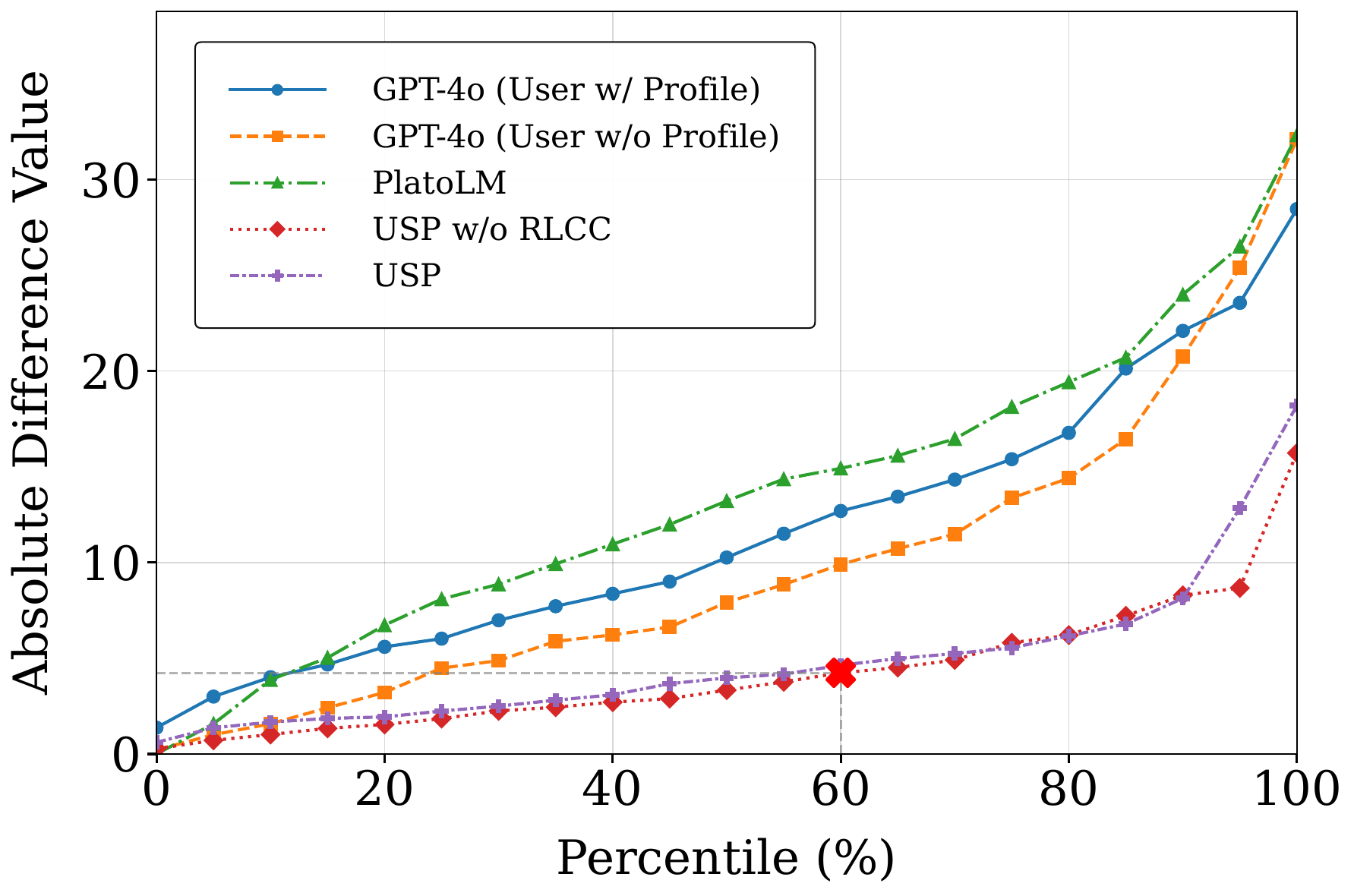}
\caption{Cumulative distribution of ADV performance comparison across different simulators, with the red cross indicating that USP (w/o RLCC) has 60\% of its samples with ADV below 5\%.}
\label{fig:target_generate_visualization}
\end{figure}

Figure~\ref{fig:target_generate_visualization} compares ADV between target and generated dialogues across simulators and percentiles. The USP series consistently demonstrate lower ADV, even in extreme cases, with 60\% of samples achieving ADV below 5\% (marked by a red cross), compared to baselines(e.g. PlatoLM, GPT-4o (User w/ Profile)) at 10\% or higher. This reflects the USP series' superior ability to generate dialogues that closely align with target conversations, effectively preserving the diversity distribution of user characteristics.

We also show that our sampling strategy outperforms random sampling by effectively capturing diverse representatives (majority and minority) in Appendix~\ref{appendix:sampling strategy effectiveness}.

\section{Analysis}
\subsection{Ablation Study}

\begin{table*}[!ht]
    \centering
    \small
    \resizebox{\textwidth}{!}{%
    \begin{tabular}{l *{9}{c}}
    \toprule
    \multirow{2}{*}{\textbf{Model Configuration}} & \multicolumn{1}{c}{\textbf{Continuity}} & \multicolumn{3}{c}{\textbf{Authenticity}} & \multicolumn{5}{c}{\textbf{Consistency}} \\
    \cmidrule(lr){2-2} \cmidrule(lr){3-5} \cmidrule(lr){6-10}
    & ESR$\downarrow$ & Sem-Sim$\uparrow$ & Style-Sim$\uparrow$ & AVA$\uparrow$ & r-DP.P$\uparrow$ &r-DP.R$\uparrow$ & r-DPC$\uparrow$ & P.Cover$\uparrow$ & SC.Score$\uparrow$ \\
    \midrule
    USAP (w/o RLCC) & 17 & 64.22 & 40.96 & 35.55 & \textbf{65.36} & 51.29 & 57.47 & \textbf{50.35} & 3.21 \\ 
    USP (w/o RLCC) & 12 & 66.17 & 40.01 & 35.68 & 53.17 & 71.88 & 61.13 & 42.63 & 3.24 \\ 
    USP (5:5) & 14 & 66.28 & 41.22 & 37.03 & 52.23 & 71.59 & 60.39 & 43.58 & \textbf{3.55} \\
    USP (8:2) & \textbf{10} & 65.39 & \textbf{46.23} & \textbf{38.77} & 56.24 & \textbf{74.38} & \textbf{64.05} & 44.08 & 3.35 \\
    USP (9:1) & 12 & \textbf{66.91} & 38.87 & 33.62 & {58.36} & 70.62 & 63.90 & {46.75} & 3.33 \\
    
    \bottomrule
    \end{tabular}
    }
    \caption{Ablation study of our USP framework.}
    \label{tab:ablation_study}
\end{table*}

We evaluate the effectiveness of the polishing step in our two-stage profile construction pipeline, which converts attributes into natural language descriptions, by comparing it to a baseline (USAP w/o RLCC) that uses only attributes without polishing. As shown in the first two rows of Table~\ref{tab:ablation_study}, the polishing step enhances generalization, improving performance across most metrics of Continuity, Authenticity, and Consistency. In contrast, relying solely on attributes leads USAP (w/o RLCC) to excessively replicate profile descriptions, resulting in a high P.Cover score (50.35) due to attributes appearing directly in the dialogue.

We also assess the relative importance of RLCC’s two rewards by testing different \(\lambda\) values in Equation~\ref{eq:reward_define}, denoted as USP(\(\lambda\): \(1 - \lambda\)). Table~\ref{tab:ablation_study} shows that \(\lambda = 0.8\) optimally balances model capabilities and dialogue consistency. Higher \(\lambda\) (0.9) sacrifices speaking style authenticity without improving r-DPC, increasing P.Cover and indicating superficial profile matching. Conversely, \(\lambda = 0.5\) preserves authentic style but lacks sufficient consistency, resulting in stagnant performance.

\subsection{Applications: Dynamic Multi-turn Evaluation For LLMs}

One application of our simulator is addressing the gap in dynamic multi-turn evaluation of LLMs. While current automatic evaluations rely on static preset questions~\cite{zheng2024judging, bai2024mt}, our user simulator can dynamically interact with the LLM over multiple rounds, adjusting based on response quality and given user traits.

We generated 300 diverse user profiles using our sampler: 100 highest-probability (majority), 100 lowest-probability (minority), and 100 random synthetic (virtual) profiles. Using these profiles, USP engages in multi-turn dialogues with the LLM. Evaluation results, based on MT-Bench~\citep{zheng2024judging} and presented in Table~\ref{tab:downstream}, show that dynamic multi-turn evaluation aligns closely with average rankings on LiveBench~\cite{livebench} and Chatbot-Arena~\cite{chiang2024chatbot}, while revealing fine-grained weaknesses of different LLMs when interacting with specific user groups. Detailed analysis is in Appendix~\ref{appendix:downstream_analysis}.

\begin{table}[!ht]
    \centering
    \resizebox{\columnwidth}{!}{%
    \begin{tabular}{@{}lccccc@{}}
    \toprule
        \multirow{2}{*}{Model Setup} & \multicolumn{3}{c}{Sampling Strategy} & \multirow{2}{*}{Avg.} & \multirow{2}{*}{Ranking in LiveBench/}  \\
    \cmidrule(lr){2-4}
    & Major & Minor & Virtual & & Chatbot-Arena\\
    \midrule
    Deepseek-v3 & \textbf{8.25} & 6.13 & \textbf{7.70} & \textbf{7.36} & 1  \\
    GPT-4o & 7.86 & \textbf{6.65} & 7.19 & 7.23 & 3 \\
    Claude-Sonnet & 7.18 & 6.61 & 7.48 & 7.09 & 2 \\
    4o-Mini & 6.84 & 5.70 & 5.52 & 6.02 & 4\\
    Claude-Haiku & 4.88 & 5.42 & 5.43 & 5.24 & 5 \\
    \bottomrule
    \end{tabular}
    }
    \caption{LLM performance across user groups.}
    \label{tab:downstream}
\end{table}

\subsection{Case Study}
\label{sec:case_study}
Table~\ref{tab:case_study} shows that the role-playing baseline GPT-4o (w/ Profile) often copies abstract profile traits verbatim to assert user identity. In contrast, our USP conveys these traits more naturally by transforming abstract concepts into concrete and coherent expressions. For instance, when the profile states “being a father of two,” GPT-4o (w/ Profile) repeats it directly, while USP implicitly reflects this by mentioning a “son” and a “daughter” in later turns. Similarly, rather than restating “likes Italian food,” USP refers to a specific dish like pasta. These examples illustrate that USP better mimics human behavior by expressing abstract, high-level traits both directly and subtly in a natural manner, likely contributing to its stronger human preference (see Table~\ref{tab:human-eval}).  Additional case studies are provided in Appendix~\ref{appendix:case_study}.

\begin{table}[ht]
\centering
\small
\begin{tabular}{@{}p{0.97\linewidth}@{}}
\toprule

\textbf{\textit{(Profile)}}  \underline{\textbf{As a father of two}}, your strong sense of family responsibility is evident... Your interest in \textbf{vegan Italian-inspired snack dishes}...  \\
\midrule
\textit{(1st turn)} \textbf{GPT-4o (w/ Profile):} \underline{\textbf{As a father of two}} who has a keen interest in ... \\
\addlinespace[0.3em]
\textit{(2nd turn)} \textbf{GPT-4o (w/ Profile):} Lately, I've been exploring new \textbf{vegan Italian-inspired snack dishes}... \\
\midrule
\textit{(4th turn)} \textbf{USP:}\underline{\textbf{\textit{My son}}} wants \textbf{\textit{pasta}} but he doesn't like tomato sauce... \\
\addlinespace[0.3em]
\textit{(5th turn)} ...\underline{\textbf{\textit{my daughter}}} says she does not feel well... She prefers \textbf{\textit{vegan}} food in \textbf{\textit{Italian style}} ... \\
\bottomrule
\end{tabular}
\caption{Case study comparing outputs from USP and the GPT-4o (w/ Profile) baseline for a sample profile. \textbf{Bold} highlights keywords explicitly copied by GPT-4o, while \textbf{\textit{bold italic}} marks USP’s implicit or fuzzy matches.}

\label{tab:case_study}
\end{table}

\section{Conclusion}

In this work, we introduce the USP framework, which integrates extracted user profiles into the user simulator by conditional SFT and RLCC. Our experimental results, validated by both automatic metrics and human evaluations, show that USP significantly outperforms role-playing simulators (e.g., GPT-4o (User w/o Profile)) and direct simulation approaches (e.g., PlatoLM) in authenticity and diversity while achieving comparable consistency at both the sentence and conversation levels. Additionally, dynamic evaluations with various LLMs across diverse demographic groups highlight USP's effectiveness in real-world scenarios. Nonetheless, a gap remains compared to real human behavior, and our future work will explore finer-grained control and multimodal simulation.

\section*{Limitations}
We acknowledge the following limitations: 1) \textbf{Scenario Applicability}: Experiments were conducted on a single dataset, with minimal validation across others to confirm broader applicability.
2) \textbf{Linguistic and Cultural Scope}: Our focus on English dialogues may limit the applicability of USP to other languages and cultural contexts.

\section*{Acknowledgments}
This research is supported by the project of Shenzhen Science and Technology Research Fund (Fundamental Research Key Project, Grant No. JCYJ20220818103001002), Shenzhen Science and Technology Program (Shenzhen Key Laboratory, Grant No. ZDSYS20230626091302006), Shenzhen Stability Science Program 2023, Shenzhen Key Lab of Multi-Modal Cognitive Computing, SRIBD Innovation Fund (Grant No. K00120240006), and Program for Guangdong Introducing Innovative and Entrepreneurial Teams, Grant No. 2023ZT10X044.


\bibliography{custom}

\appendix

\section{Dataset Construction}

\subsection{Preprocessing}
\label{appendix:preprocess}
Our dataset preprocessing follows the method outlined in PlatoLM~\cite{KongFWJW24}, which includes the removal of non-English content, filtering of toxic data, elimination of exact duplicates at the dialogue level, and segmentation of conversations into maximum-length token sequences. To maintain discourse integrity, truncated dialogues are ensured to start with the assistant's turn, preserving context consistency and coherence.

\subsection{Profile Dataset}
\label{appendix:profile_dataset}
As detailed in Section~\ref{sec:profile_construction}, we classify attributes into three types: scene-consistent, scene-related, and deep intrinsic characteristics. For each, we use specific prompts (Figures~\ref{fig:scene_consistent_extraction}, \ref{fig:scene_related_extraction}, and \ref{fig:personality_extraction}), with metric definitions based on~\cite{abs-2406-13960} and Big Five traits per~\cite{gosling2003very}.

We then concatenate these attributes, remove invalid entries, and shuffle their order to prevent positional bias. The combined attributes are rephrased using GPT-4o with the prompt in Figure~\ref{fig:rephrase_prompt}, producing automatically labeled profiles. The LMSYS-USP dataset averages 1,149 tokens in training, 1,295 in validation, 1,438 in testing, and 231 tokens per profile.

\begin{figure}[h!]
    \centering
    \includegraphics[width=\columnwidth]{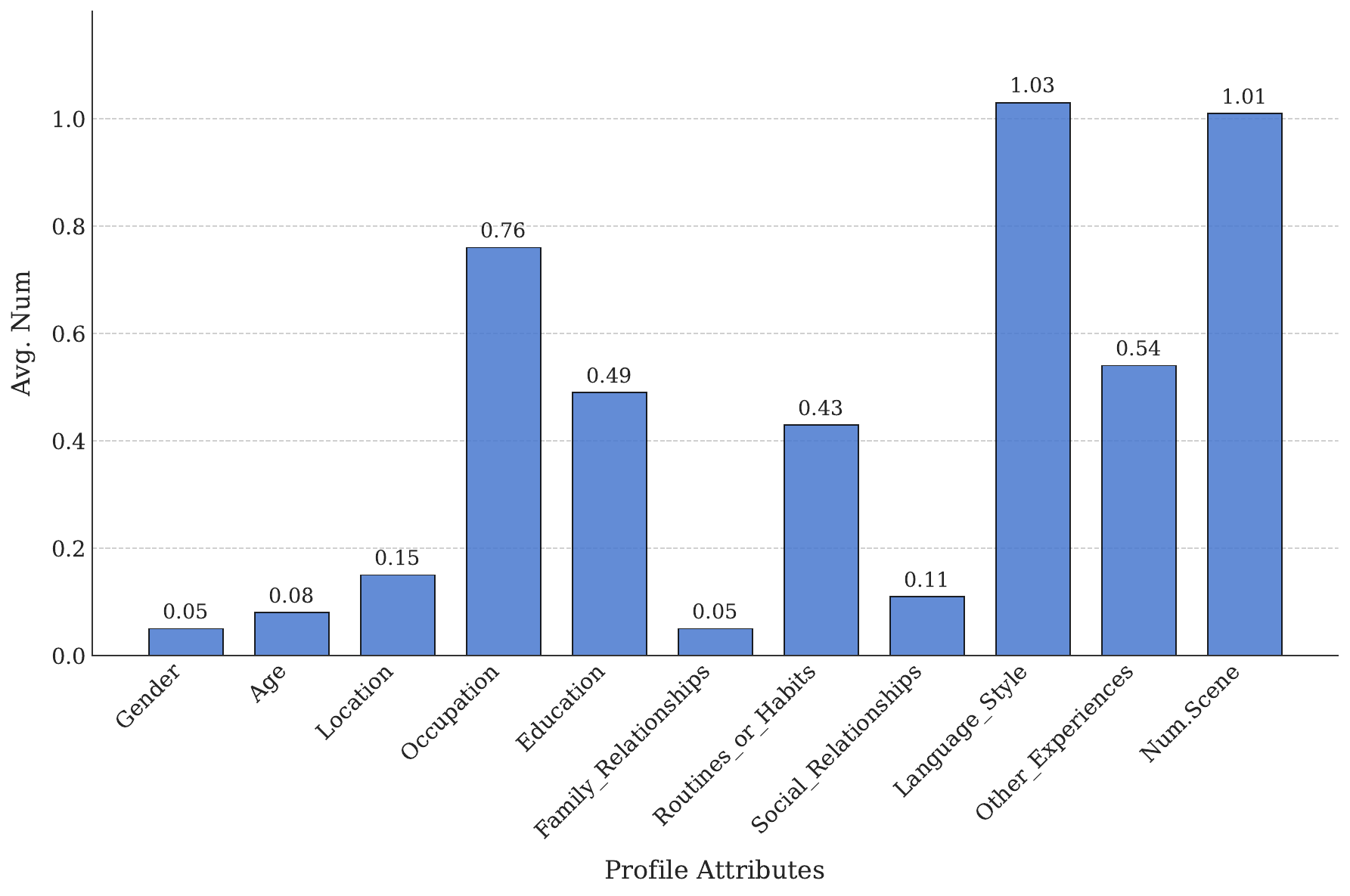}
    \caption{Frequency of values for each attribute of objective facts in the attribute extraction process.}
    \label{fig:object_fact_stats}
\end{figure}

We also measured the frequency of each attribute value, defined as the average number of distinct values per sample, to assess attribute prevalence. Statistics for objective facts are shown in Figure~\ref{fig:object_fact_stats}. For subjective traits, we focused on the Big Five traits only when scores were significantly high or low, excluding moderate scores as they reflect average human behavior~\cite{MoonAKSSBC24} and were omitted from the subsequent polishing step.

\begin{table}[h!]
    \centering
    \label{tab:subject_characters}
    \resizebox{\columnwidth}{!}{%
    \begin{tabular}{lcc}
        \toprule
        \textbf{Attribute} & \textbf{High Rate (\%)} & \textbf{Low Rate (\%)} \\
        \midrule
        Conscientiousness & 78.07 & 7.53 \\
        Agreeableness & 6.45 & 14.98 \\
        Extraversion & 4.08 & 14.15 \\
        Openness & 58.77 & 5.30 \\
        Neuroticism & 2.04 & 10.12 \\
        \bottomrule
    \end{tabular}
    }
    \caption{Summary of extracted subjective attribute statistics.}
\end{table}

\subsection{Resource Consumption in Implementation}

Attribute extraction using the GPT-4o API costs ~\$0.003 per attribute type, or ~\$0.01 per sample for three types. For 94,000 samples, the extraction costs ~\$940. Rewriting attributes into profiles adds ~\$0.05 per sample, resulting in a total dataset construction cost of ~\$1,400.

\section{Implement Detail}
\subsection{Trainable Model Setup}
\label{appendix:model_setup}

For PlatoLM, we base it on the LaMA-3-8B-Base architecture. Following~\cite{KongFWJW24}, the system prompt is:
\texttt{"A chat between a curious human and an artificial intelligence assistant. The human can ask follow-up or new questions without prior context."}
We fine-tune using four A100 40GB GPUs for 3 epochs, taking about two days.

The AI detection model uses Longformer~\cite{beltagy2020longformer} trained on our dataset per~\cite{cheng2025beyond}. User utterances are labeled as human, and assistant utterances as AI. Training runs for 3 epochs on dual RTX 3090 GPUs, taking three days.

The profile generator is fine-tuned from LLaMA-3-8B-Instruct~\cite{llama3modelcard} on our curated profile dataset, effectively distilling GPT-4o’s two-stage profile generation. Training uses four A100 40GB GPUs for 3 epochs, lasting two days.

\subsection{Train-Free Model Setup}
We use two simulator types: (1) A response model, e.g., GPT-4o (User with Profile), role-playing as a user simulator, unable to initiate conversations. The user profile is embedded in the system prompt, and the initial query is generated by asking, "What will you say to start the conversation?" to obtain the user's opening query. (2) A user simulator, e.g., PlatoLM, directly generates user utterances from a seed prompt without additional steps.

\subsection{Human Evaluation}
\label{appendix:huamn_eval}

All annotators we recruited were based on two criteria: (1) an IELTS score of 6.5 or higher for sufficient English proficiency, and (2) a Computer Science background with research experience or foundational knowledge in dialogue systems.

\subsubsection{Profile Evaluation} 
Two annotators (one undergraduate, one master’s student) rated extracted profiles on a 1–5 scale based on dialogues, assessing: (1) accuracy of objective facts (precision without hallucinations), (2) completeness (no significant omissions), and (3) reasonableness of subjective descriptions (rational, unbiased, justified). Results in Table~\ref{tab:profile_human_eval} indicate moderate to high annotator agreement. 

\begin{table}[h!]
   \centering
   \resizebox{\columnwidth}{!}{%
   \begin{tabular}{lcccc}
       \toprule
       Dataset & Profile Source & Objective Facts & Subjective Characters & Naturalness \\
       \midrule
       LMSYS-USP & GPT-4o & 4.64 & 4.19 & 4.66 \\
       \bottomrule
   \end{tabular}
   }
    \caption{Human evaluation results for profile quality across three aspects: Objective Facts, Subjective Characters, and Naturalness, with 1-point agreement rates of 89.2\%, 74.3\%, and 88.4\% respectively.}
   \label{tab:profile_human_eval}
\end{table}

\subsubsection{Dialogue Evaluation}
We recruited eight annotators—comprising two undergraduates, five postgraduates, and one postdoctoral researcher—to evaluate conversation-level results. This diverse academic representation ensured a broad range of expertise. Annotators assessed dialogues using two key criteria: authenticity and consistency. For authenticity, they compared user utterances against a reference dialogue, focusing on style, semantics, and quality. For consistency, annotators evaluated user utterances solely based on the provided user profile, considering accuracy, completeness, and quality. These definitions align with prior work~\cite{cheng-etal-2024-dialogues, abs-2406-13960}, with detailed guidelines provided in Figure~\ref{fig:human_eval_authenticity} and Figure~\ref{fig:human_eval_consistency}.

Eight annotators were randomly paired into four groups, each independently evaluating a randomly assigned dialogue sample. Each group reviewed 75 examples across three baselines (100 examples each). To reduce position bias and prior exposure, dialogue pair assignments and their presentation order were randomized. 

\begin{figure}[t]
    \centering
    \includegraphics[width=\linewidth]{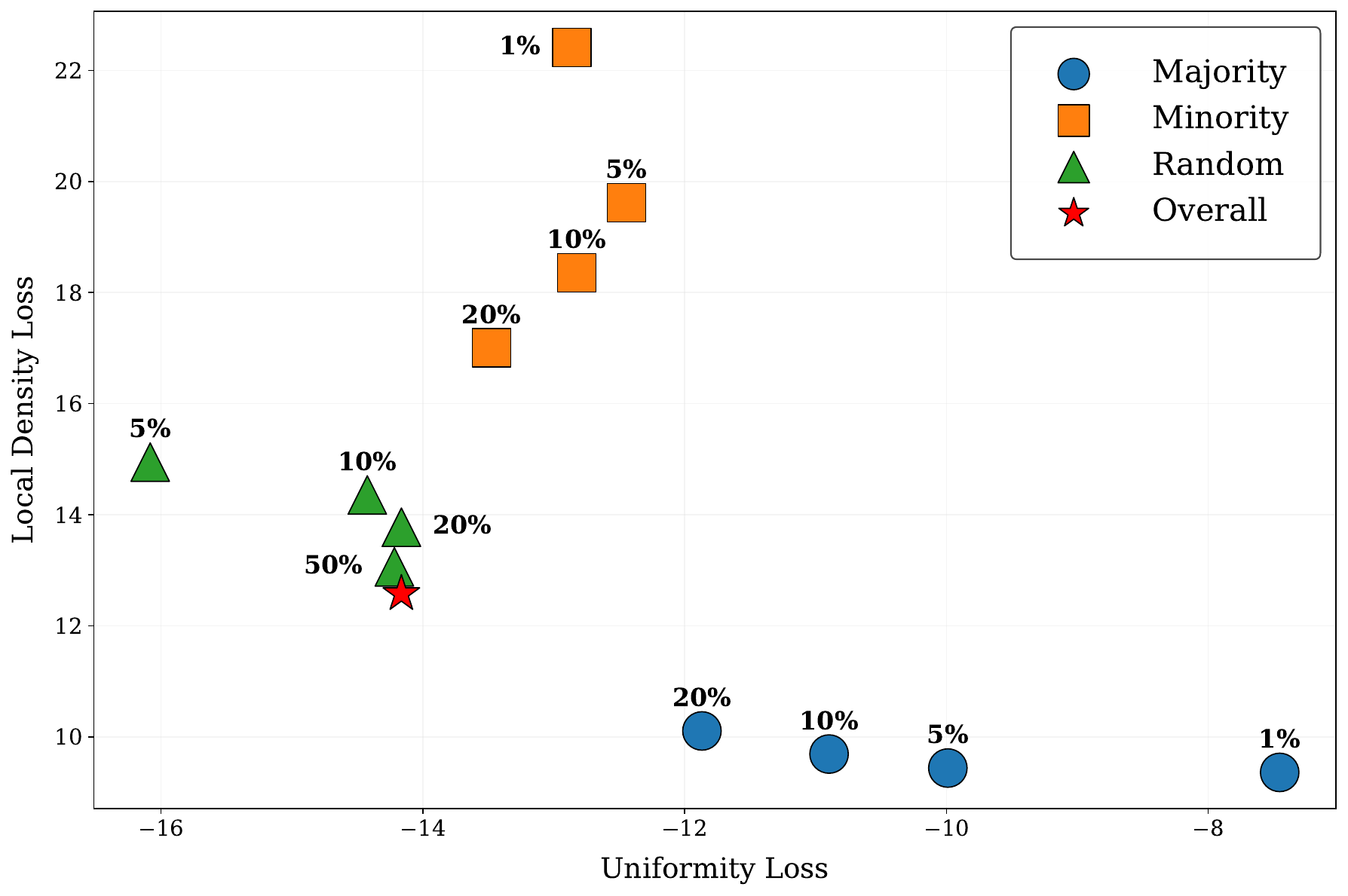}
    \caption{Distribution of different sampling strategies.}
    \label{fig:density_analysis}
\end{figure}

\section{Further Analysis}
\subsection{Sampling strategy effectiveness}
\label{appendix:sampling strategy effectiveness}

To evaluate our density sampler, we use two complementary metrics: Local Density Loss (LDL)\cite{doi:10.1126/science.1242072} to assess structure preservation, and Uniformity Loss\cite{wang2020hypersphere} to measure global coverage. Lower LDL indicates tighter local clustering, preserving natural profile structures, while lower Uniformity Loss reflects better global coverage with realistic distributions.

Guided by GKDE density estimates, we apply two strategies: sampling high-density regions to capture majority patterns and weighting low-density regions to cover minority cases. Figure \ref{fig:density_analysis} shows how this approach balances distribution preservation and targeted sampling. Moving along the uniformity loss axis reveals a shift from majority samples (blue circles), which excel in low LDL and high uniformity regions, to minority samples (orange squares), which occupy higher LDL areas with moderate uniformity to capture diversity. Sampling percentages progress steadily for both, indicating controlled behavior, while random sampling (green triangles) displays scattered patterns, confirming our method’s reliability. The overall performance (red star) highlights a successful balance between preservation and targeted sampling.

\subsection{Case Study}
\label{appendix:case_study}
To evaluate USP’s performance on consistency and authenticity, we present two dialogues generated via interactive conversations with GPT-4o. Figure \ref{fig:case_study_authenticity} assesses authenticity by comparing USP with reference dialogues and other baselines, while Figure \ref{fig:case_study_consistency} examines profile consistency across profile-based models.

USP captures stylistic nuances—such as the consistent use of lowercase “i” and concise questioning—and maintains strong semantic alignment with target conversations. In contrast, PlatoLM diverges from the dialogue flow by the fourth turn, and GPT-4o (User w/ Profile) falls into repetitive praise.

On consistency, USP effectively integrates objective profile details and subjective traits, demonstrating strong generalization to unseen user profiles.

\subsection{Downstream analysis}
\label{appendix:downstream_analysis}

\begin{figure*}[htp]
    \centering
    \subfloat[Performance on the majority group]{
    \label{fig:major_model_performance}
    \includegraphics[width=0.32\linewidth]{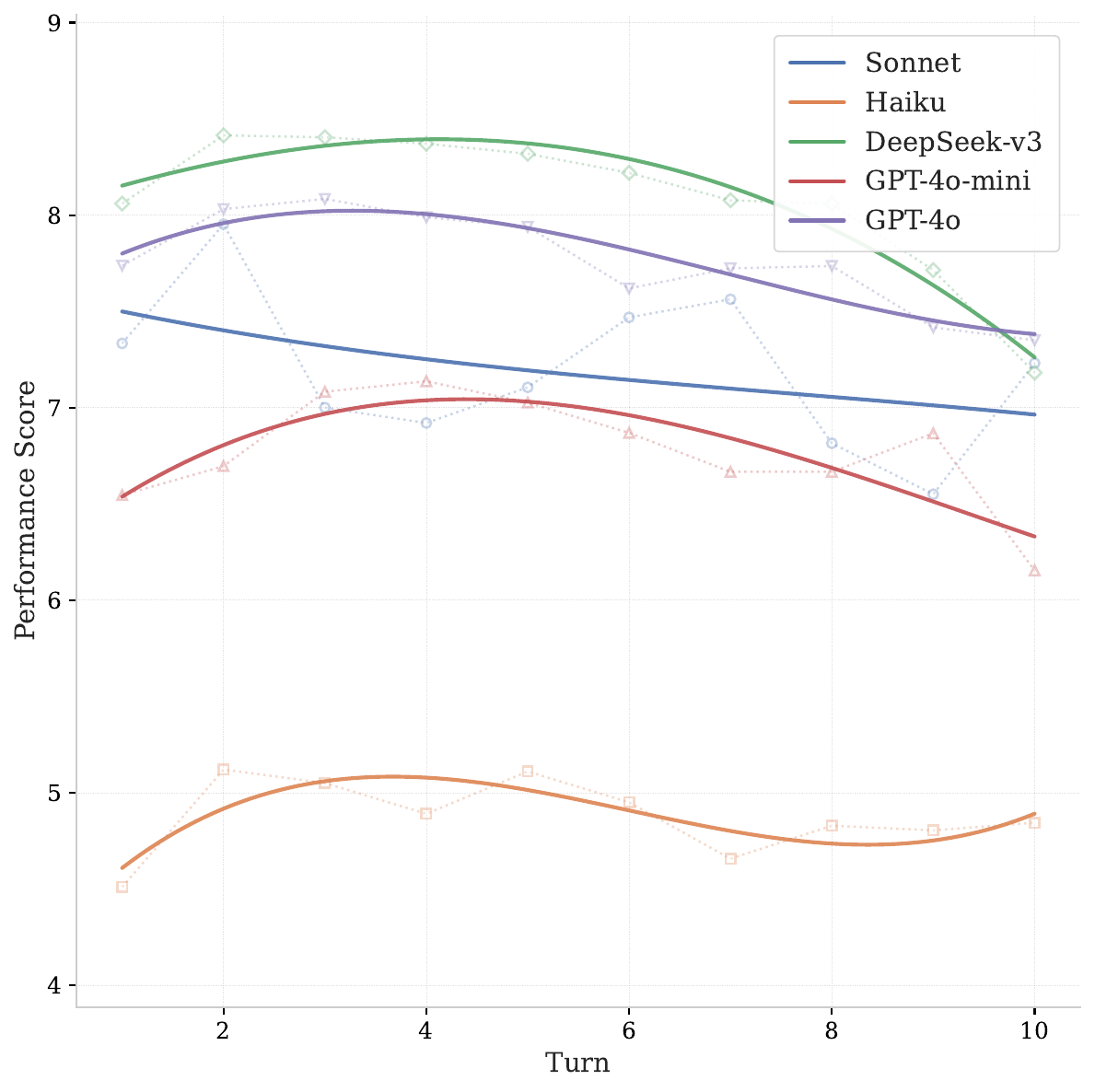}\hfill
    }
    \subfloat[Performance on the minority group]{
    \label{fig:minor_model_performance}
    \includegraphics[width=0.32\linewidth]{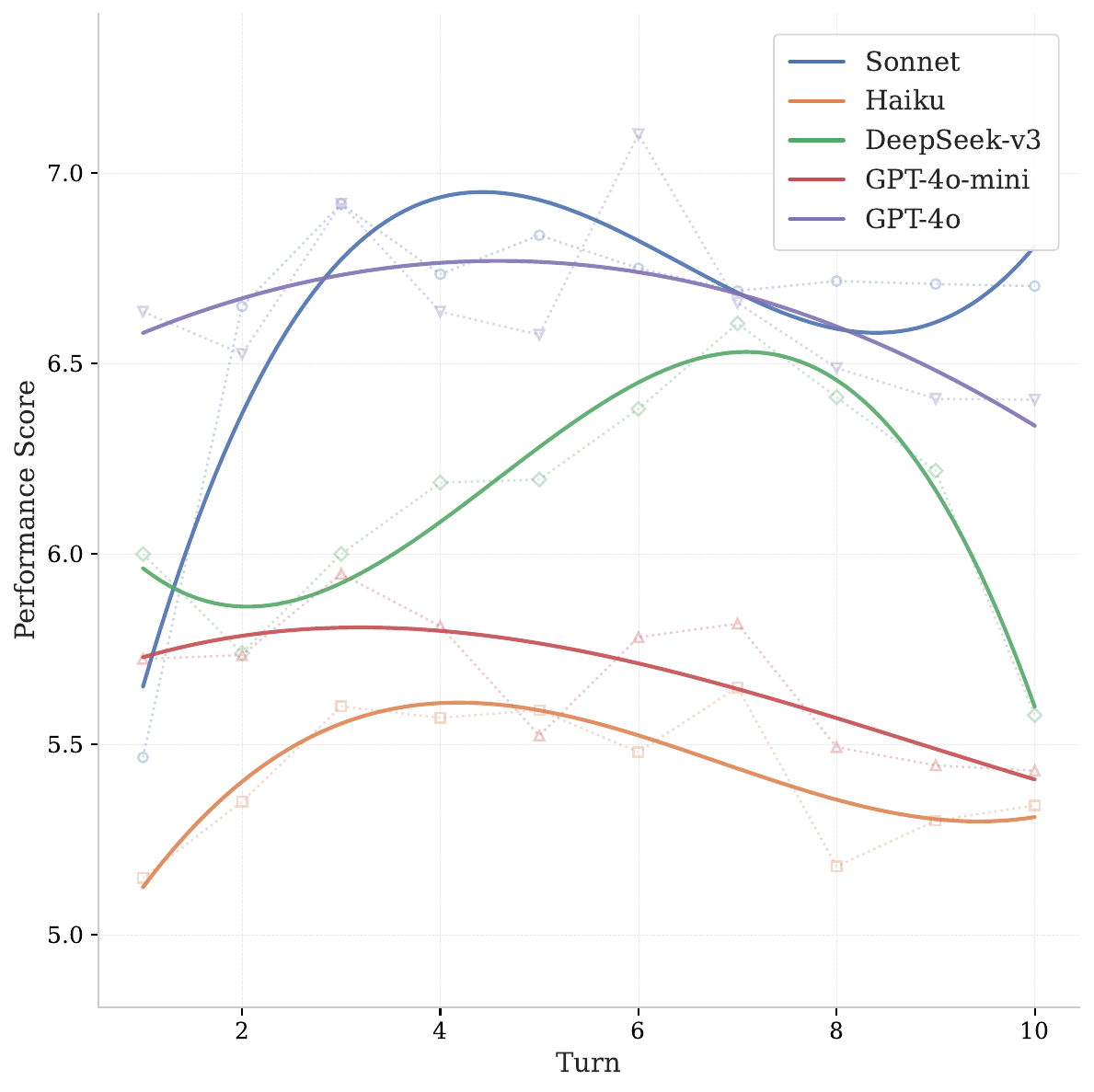}\hfill
    }
    \subfloat[Performance on the virtual group]{
    \label{fig:virtual_model_performance}
    \includegraphics[width=0.32\linewidth]{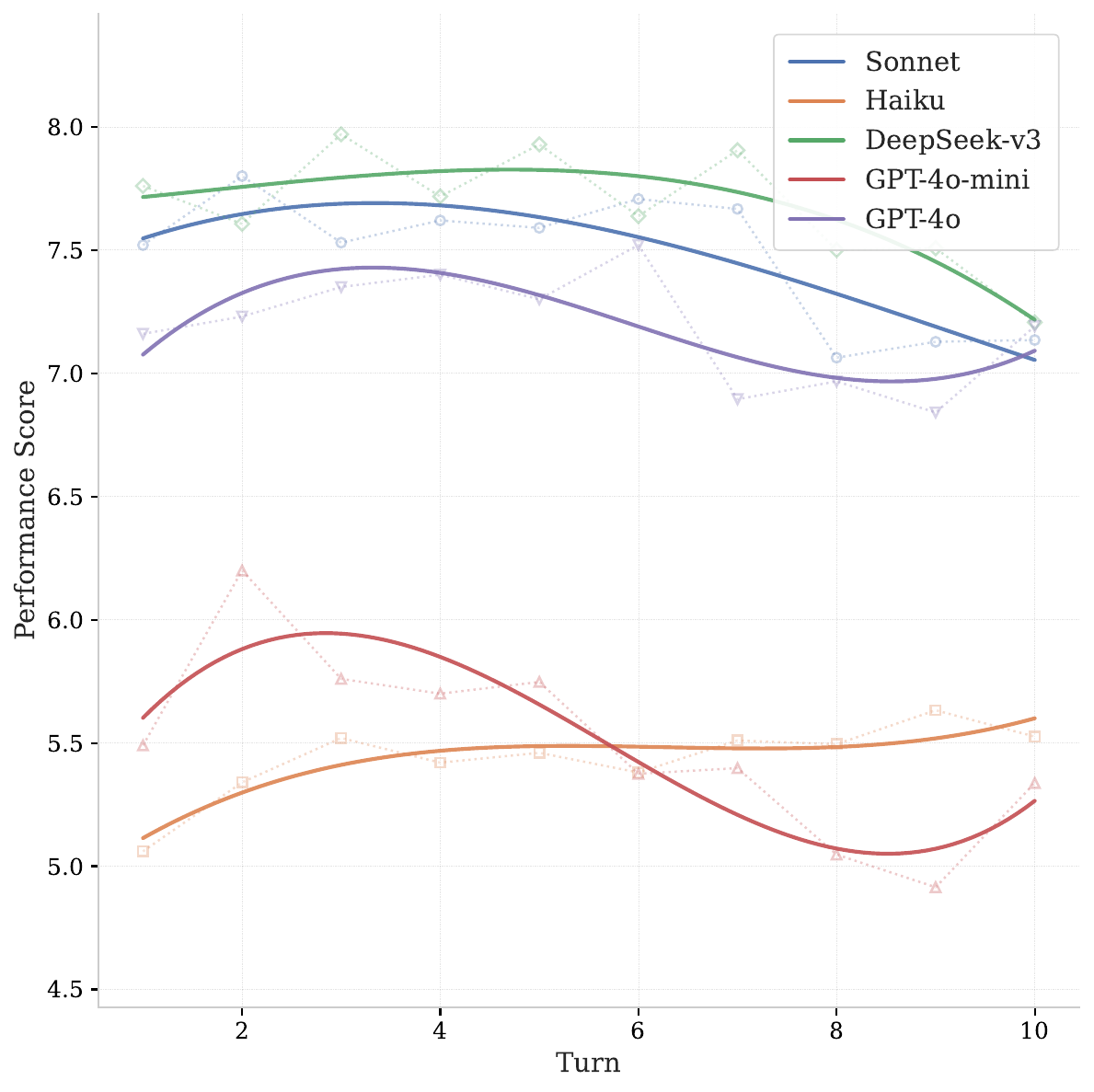}\hfill
    }
    
    \caption{Performance trends of different response models across dialogue turns for various demographic groups.}
    \label{fig:downstream}
\end{figure*}

Our analysis of performance trends across dialogue turns for mainstream LLMs with different demographic groups reveals four key findings, as illustrated in Figure~\ref{fig:downstream}:
(1) While LLMs demonstrate robust performance with the majority demographics, they show notably decreased overall effectiveness when interacting with minority groups, highlighting limitations in personalization capabilities;
(2) The models maintain reasonable performance with virtual groups, suggesting effective generalization abilities beyond real-world demographics;
(3)  Instruction-following capability gradually declines as dialogue turns increase, aligning with observations from previous studies~\cite{kwan-etal-2024-mt, maharana2024evaluating};
(4) The pronounced performance volatility across dialogue turns for minority groups underscores the need for enhanced capabilities in processing and responding to less common interaction patterns.

\section{Prompt Templates}
\label{appendix:prompt_design}
\begin{figure*}[t] 
\begin{tcolorbox}[colback=white, colframe=black, title=Human Evaluation Guidelines for Authenticity]
1. \textbf{Task Description:}
 Please choose which user in the two test conversations is more similar to the reference conversation being spoken by the same person.
 
2. \textbf{Evaluation Criteria:}
     \begin{itemize}
         \item \textbf{Semantic Similarity}: Measure the thematic consistency and discourse coherence between the generated user utterance and the target user utterance. Preference should be given to the utterance that more accurately reflects the semantic content of the target.
         \item \textbf{Stylistic Parity}: Analyze whether the generated user utterance matches the style of the target user utterance, including its tone, vocabulary, and grammatical structure. The utterance that aligns more closely with the stylistic elements of the target should be favored.
         \item \textbf{Quality}: Examine the fluency and logical coherence of the user utterance, focusing on the linguistic and logical smoothness of the user utterance. The more coherent and fluent utterances should be chosen.
     \end{itemize}
\end{tcolorbox}
\caption{Human evaluation guidelines for authenticity.}
\label{fig:human_eval_authenticity}
\end{figure*}

\begin{figure*}[t] 
\begin{tcolorbox}[colback=white, colframe=black, title=Human Evaluation Guidelines for Consistency]
1. \textbf{Task Description:} 

     Select the dialogue that contains the most appropriate \textbf{user utterance} from the two given generated dialogues based on the target user profile. 

2. \textbf{Evaluation Criteria:}

     \begin{itemize}
    \item \textbf{Persona Reflection}: How well does the utterance capture the profile’s themes, tone, and style? Prefer utterances that align closely with the profile’s characteristics.
    \item \textbf{Comprehensiveness}: Does the utterance integrate both objective facts and subjective traits? Favor broader, more complete reflections over narrow ones.
    \item \textbf{Quality}: Evaluate fluency, coherence, and naturalness. Prefer utterances that read smoothly, are logically consistent, and resemble genuine human conversation.
     \end{itemize}

\end{tcolorbox}
\caption{Human evaluation guidelines for consistency.}
\label{fig:human_eval_consistency}
\end{figure*}

\begin{figure*}[t] 
\begin{tcolorbox}[colback=white, colframe=black, title=Case Study on Authenticity of Simulated Dialogue by Different User Simulators]
{\bf GPT-4o (User w/ Profile) Simulated Dialogue:} \\
{\bf GPT-4o (User w/ Profile): } Hey there! I'm really excited to dive into some AI discussions with you...\\
\textbf{Assistant:} Hello! I'm thrilled to chat about AI with you... \\
{\bf GPT-4o (User w/ Profile): } It's great to dive into these with someone who shares such a passion...\\
\textbf{Assistant:} I'm glad to hear that you're as passionate about AI as I am! Your interests...\\
{\bf GPT-4o (User w/ Profile): } I'd be delighted to guide you in translating these advanced AI concepts...\\
\textbf{Assistant:} I'm excited to help guide your ideas into actionable solutions ...\\
{\bf GPT-4o (User w/ Profile): } I'm thrilled at the prospect of collaborating to develop solutions...\\
\textbf{Assistant:} I'm excited at the prospect of working together to make these AI strategies...
\tcbline
{\bf PlatoLM Simulated Dialogue:} \\
{\bf PlatoLM: } how to train a model like you\\
\textbf{Assistant:} I am a large language model... preprocessing step involves tokenizing the text...\\
{\bf PlatoLM: } what do you mean by tasks such as tokenizing text... explain these terms please?\\
\textbf{Assistant:} Certainly! These terms refer to common preprocessing steps...\\
{\bf PlatoLM: } so what does the final product look like when we finish tokenizing?\\
\textbf{Assistant:} Absolutely! Let's go through a brief example step by step...\\
{\bf PlatoLM: } who painted NAME\_27? Context knowledge: \#title NAME\_27 ...\\
\textbf{Assistant:} Based on the provided context, the artist referred to as...
\tcbline
{\bf USP Simulated Dialogue:} \\
{\bf USP: } hello who am i talking to?\\
\textbf{Assistant:} Hello! You're speaking with ChatGPT...\\
{\bf USP: } are you vicuna?\\
\textbf{Assistant:} No, I'm not Vicuna. I'm ChatGPT, created by OpenAI...\\
{\bf USP: }  what do you know about it?\\
\textbf{Assistant:} Vicuna is an AI language model developed by a different research group...\\
{\bf USP: } can you tell me how i could train and deploy my own ai model?\\
\textbf{Assistant:} Training and deploying your own AI model involves several key steps...
\tcbline
{\bf Reference Dialogue:} \\
{\bf User: }  how to train a model like you\\
\textbf{Assistant:} I am a large language model, which means that I was trained on...\\
{\bf User: }  what is your base model\\
\textbf{Assistant:} I am a large language model called GPT...\\
{\bf User: }   isn't you vicuna\\
\textbf{Assistant:}  I am not aware of any information or research about a model called "vicuna"...\\
{\bf User: }  what hardware do i need to run you\\
\textbf{Assistant:} I am a large language model called GPT, which was trained on...\\

\end{tcolorbox}
\caption{Case study comparing USP with other user simulators over the first four of ten dialogue turns. USP and GPT-4o (User w/ Profile) rely solely on the given profile, while PlatoLM uses the first-turn golden context. All simulators interact with GPT-4o, aiming to reconstruct the reference dialogue shown below.}

\label{fig:case_study_authenticity}
\end{figure*}

\begin{figure*}[t] 

\begin{tcolorbox}[colback=white, colframe=black, title=Case Study on Consistency of Simulated Dialogues across Different User Simulators]
{\bf Profile:} You are engaging in a conversation with an AI assistant. Your profile is:
\\
You have a passion for engaging in activities that challenge your vocabulary and word formation skills. One of your favorite pastimes is diving into puzzle games that offer a thrilling test of language prowess. You often find yourself thinking of five-letter words with tricky placements, such as words containing 'A' in the fourth position, or concocting ones with 'R' in the second spot and 'T' at the end. This aligns with your appreciation for linguistic puzzles and an enduring enjoyment of wordplay, reflecting your cognitive curiosity to enhancing your language skills. \\

Your vibrant personality shines through a playful, engaging language style, crafting dynamic game scenarios with structured questions. You tackle intellectual challenges with vivid imagination and meticulous problem-solving. Highly organized and detail-oriented, you excel in complex tasks. Embracing a reserved lifestyle, you find joy in solitary, introspective pursuits, diving into intricate intellectual endeavors with enthusiasm and precision. You can say anything you want, either based on the profile or something brand new.
\tcbline
{\bf GPT-4o (User w/ Profile) Simulated Dialogue:} \\
{\bf GPT-4o (User w/ Profile): } How about a playful journey into the world of word puzzles? Let's start with...\\
\textbf{Assistant:} Sure! How about the word "brave"? The letter 'A' is in the fourth position.  \\
{\bf GPT-4o (User w/ Profile):} Wonderful! "Brave" is intriguing. Now, let's spice it up...\\
\textbf{Assistant:} Absolutely! The word "crest" fits your criteria, with 'R'...\\
{\bf GPT-4o (User w/ Profile): } Splendid! "Crest" evokes such a vivid image; I love it....\\
\textbf{Assistant:} Of course! The word "shine" fits your criteria, with 'S' as the first letter ...\\
{\bf GPT-4o (User w/ Profile): } Brilliant choice! "Shine" has such a radiant quality to it...\\
\textbf{Assistant:} I'm glad you're enjoying it! How about this: I'll give you a word puzzle...
\tcbline
{\bf USP Simulated Dialogue:} \\
{\bf USP: } Let's play a game about 5 letter words. I'll ask you some questions and try to guess the answer. Can we start?\\
\textbf{Assistant:} Sure, I'd be happy to play a game with you! Let's get started!\\
{\bf USP: } Question: What is a 5-letter word which has "A" in its 4th place? Answer:\\
\textbf{Assistant:} Is the word you are thinking of "Polar"? \\
{\bf USP: }  Nope What about "gnaw"?\\
\textbf{Assistant:}  "GNW" is a four-letter word. Let's think of another five-letter word with "A" in the...\\
{\bf USP: } Next Question: Give me an example of a 6-letter word with R at second place and T at sixth place.\\
\textbf{Assistant:} Sure! A six-letter word with "R" in the second position and "T" at the sixth position...\\
{\bf USP: } Great! Now give me one example of a 7-letter word with S at 3rd place and H at 7th place.\\
\textbf{Assistant:} Sure! One example of a seven-letter word with "S" in the third place and "H" in ..."
\tcbline
{\bf Reference Dialogue:} \\
{\bf User:}  You are a very intelligent puzzle solver. We are going to play a game...Question: What is a 5 letter word which has "A" in 4th position? Answer:\\
\textbf{Assistant:} Panda.\\
{\bf User:}  Question: Suggest a 5 letter word with R in second position and T in fifth position. Answer:\\
\textbf{Assistant:} Graft.
\end{tcolorbox}
\caption{Case study comparing user simulators over the first four turns of a 10-turn dialogue. USP and other simulators interact with GPT-4o using only the provided profile, targeting the reference dialogue for reconstruction.}
\label{fig:case_study_consistency}
\end{figure*}

\begin{figure*}[t]
\begin{tcolorbox}[colback=white, colframe=black, title=Prompt for Extracting Scene-Consistent Attributes]
\textbf{User Persona Analysis Task}

\textbf{Objective} \\
The primary goal of this task is to analyze user utterances in-depth and accurately extract key persona attributes based on both direct and implicit cues. These attributes should be categorized into distinct fields, with any missing or unclear details left blank. 

\textbf{Field Descriptions:} 
\texttt{\{\{Field\}\}: \{\{definition\}\}}

\textbf{Guidelines}
\begin{enumerate}
    \item Carefully examine each user utterance to extract relevant persona traits. Consider both direct statements and implicit clues.
    \item Ensure that the extracted attributes are specific and directly relevant to the user's utterances. Avoid vague or generalized descriptions unless explicitly supported by the text.
    \item Pay attention to distinctive communication styles (e.g., formal or casual tone, frequent use of specific words or phrases) to capture the user's unique way of communicating.
\end{enumerate}

\textbf{Example}

User Utterances:
\begin{verbatim}
[User]: Given an array of integers nums and an integer target, return indices of 
the two numbers such that they add up to target in Python...
[Assitant]: To solve the problem...
[User]: Thanks
\end{verbatim}

Expected Output:
\begin{verbatim}
{
    "gender": [],
    "age": [],
    "location": [],
    "occupation": [
        "Likely a beginner programmer or student studying computer science, 
        as evidenced by the simple coding problem in Python."
    ],
    "education": [
        "Possibly a student in computer science or a related field, 
        at an early stage in learning programming, specifically Python."
    ],
    "family_relationships": [],
    "routines_or_habits": [],
    "social_relationships": [],
    "language_style": [
        "Concise and task-oriented",
        "Polite response 'Thanks' after getting satisfactory answer"
    ],
    "other_experiences": []
}
\end{verbatim}
\end{tcolorbox}
\caption{Prompt for extracting scene-consistent attributes.}
\label{fig:scene_consistent_extraction}
\end{figure*}

\begin{figure*}[t]
\begin{tcolorbox}[colback=white, colframe=black, title=Prompt for Extracting Scene-Related Attributes]
\textbf{User Persona Analysis Task}

\textbf{Objective} \\
The goal of this task is to analyze multi-turn user utterances within a conversation with an assistant and extract key elements such as the primary goals and specific task descriptions. Each extracted detail should be as specific as possible, reflecting the user's context, objectives, and scenario.

\textbf{Output Format} \\
The extraction should be presented in a structured JSON format, as shown below:

\begin{verbatim}
{
    "scenarios": [
        {
            "goals_or_plans": "<List of User's goals or plans>",
            "task_details": "<List of specific tasks summary made by the user>"
        },
        ...
    ]
}
\end{verbatim}

\textbf{Field Descriptions:}
\begin{itemize}
    \item \textbf{goals\_or\_plans}: User's short-term or long-term objectives, derived from explicit statements or inferred from the overall conversation. If no explicit goals are stated, infer them from the main topics discussed.
    \item \textbf{task\_details}: Specific tasks, actions, or requests made by the user. Each task should be concisely summarized with specifics. If there are multiple tasks, list each separately.
\end{itemize}

\textbf{Example} \\
User Utterances:
\begin{verbatim}
[User]: Summarize: Harry Potter is a fictional character in Harry Potter series...
[Assitant]: Harry Potter is a fictional character...
[User]: Write an email inquiring about coursework...
\end{verbatim}

\begin{verbatim}
{
    "scenarios": [
        {
            "goals_or_plans": "Aiming to gain a deeper understanding of the Harry
                        Potter series, possibly for academic or personal enrichment.",
            "task_details": [
                "Summarizing introductory content about the Harry Potter character."
            ]
        },
        {
            "goals_or_plans": "Looking to improve professional communication skills.",
            "task_details": [
                "Writing an email to inquire about coursework."
            ]
        }
    ]
}
\end{verbatim}
\end{tcolorbox}
\caption{Prompt for extracting scene-related attributes.}
\label{fig:scene_related_extraction}
\end{figure*}

\begin{figure*}[t]
\begin{tcolorbox}[colback=white, colframe=black, title=Prompt for Extracting Big Five Personality Traits]
\textbf{Task:}

\text{Analyze the provided dialogue to assess the user's personality traits across 5 personality dimensions.}
\text{Focus exclusively on the user's characteristics, disregarding any information related to others}
\text{individuals, unless it directly impacts the user.}

\text{For each personality trait:}
\begin{enumerate}
    \item \text{Assign each dimension independently with \textbf{score}: "High", "Low," or "Inconclusive."}
    \item \text{Provide \textbf{conclusion}: A high-level description, with concise supporting details.}
    \item \text{Provide \textbf{reason}: Justify the assigned score with specific evidence from the dialogue.}
    \item \text{Mark traits as \textbf{Inconclusive} only when no clear evidence exists.}
\end{enumerate}

\textbf{Personality Trait Defination:}

\texttt{\{\{metric\}\}: \{\{definition\}\}} \\
\textbf{Format:}
\begin{verbatim}
{
    "Trait Name": {
        "score": "High/Low/Inconclusive",
        "conclusion": "The user is a [trait descriptor] person...",
        "reason": "Explanation referencing specific dialogue evidence."
    },
    ...
}
\end{verbatim}

\textbf{Example:}
\begin{verbatim}
[User]: "She is my age, in a homeless women's shelter, living under very poor
conditions. She is a mental health client, but the treatment team seems to
... Her background is similar to mine, and I cannot abandon her."
\end{verbatim}

\textbf{Detected Personality Traits:}
\begin{verbatim}
{
  "Conscientiousness": {
    "score": "High",
    "conclusion": "The user is a conscientious person who demonstrates a
    sense of duty and commitment.",
    "reason": "The user expresses a strong sense of responsibility ..."
  },
  "Agreeableness": {
    "score": "High",
    "conclusion": "The user is an empathetic and compassionate person who
    values relationships.",
    "reason": "The user shows care and concern for their cousin's well-being..."
  },
  "Extraversion": {
    "score": "Inconclusive"
  },
  ...
}
\end{verbatim}
\end{tcolorbox}
\caption{Prompt for extracting deep intrinsic characteristics.}
\label{fig:personality_extraction}
\end{figure*}

\begin{figure*}[t]
\begin{tcolorbox}[colback=white, colframe=black, title=Prompt for Rephrasing Attributes into Natural Descriptions]
\textbf{Narrative Generation Objective} \\
Rephrase the provided key-value pairs into a natural, coherent second-person description.

\textbf{Core Requirements}
\begin{enumerate}
    \item \textbf{Perspective}: Use second-person perspective ("you", "your").
    \item \textbf{Structure}: Two paragraphs:
    \begin{itemize}
        \item First paragraph: Present objective facts.
        \item Second paragraph: Describe subjective characteristics.
    \end{itemize}
    \item \textbf{Key Principles}
    \begin{itemize}
        \item Accurately represent \textbf{all} provided details.
        \item Ensure the language flows naturally, remains engaging, and avoids redundancy.
        \item Focus on clear and seamless transitions between ideas.
    \end{itemize}
\end{enumerate}

\textbf{Output Expectations}
\begin{itemize}
    \item \textbf{Objective Facts}:  
    \begin{itemize}
        \item Convert the key-value pairs into a clear and natural description without over-explaining or adding unnecessary details.
        \item Ensure each scenario is logically connected and key information is presented effectively.
    \end{itemize}
    \item \textbf{Subjective Characteristics}: 
    \begin{itemize}
        \item Avoid vague terms like "high perfectionism" or "moderate emotional stability." Use vivid, descriptive language to bring these traits to life.
    \end{itemize}
\end{itemize}

\end{tcolorbox}
\caption{Prompt for rephrasing attributes into natural descriptions for profile generation.}
\label{fig:rephrase_prompt}
\end{figure*}

\begin{figure*}[t]
\begin{tcolorbox}[colback=white, colframe=black, title=Prompt for NLI Score of Profile Precision(DP.P) Based on Given Dialogue]
\textbf{Role} \\
You are an expert in evaluating the \textbf{consistency} between a given \textbf{user profile (Source)} and \textbf{the user’s utterance (Target)}. Your task is to assess whether the \textbf{Target} aligns with, contradicts, or is ambiguous in relation to the \textbf{Source}.

\textbf{Task Instructions:} \\
For each \textbf{Source-Target} pair, determine the relationship using the following scoring criteria:
\begin{itemize}
    \item \textbf{Score 1}: The Target is consistent with the Source (the interpretation can be inferred from the Source).
    \item \textbf{Score -1}: The Target conflicts with the Source (the interpretation contradicts the Source).
    \item \textbf{Score 0}: The relationship is unclear or ambiguous (insufficient evidence to infer consistency or contradiction).
\end{itemize}

\textbf{Output Format:} \\
Provide your result in the following JSON format:
\begin{verbatim}
{
  "score": <score>, 
  "reason": "<concise explanation of the reasoning>"
}
\end{verbatim}

\textbf{Example:} \\
\textbf{Source}: You are interested in dataset-related details. \\
\textbf{Target}: [User]: Show me how to implement a toy version of a relational database.

\textbf{Output}:
\begin{verbatim}
{
  "score": 1, 
  "reason": "The request for implementing a relational database suggests an 
  interest in data structures and datasets, which aligns with the Source."
}
\end{verbatim}

\textbf{Guidelines:}
\begin{enumerate}
    \item \textbf{Contextual Inference}: Do not evaluate the Target \textbf{in isolation}. Instead, determine its logical relationship to the Source, considering both explicit statements and reasonable implications.
    \item \textbf{Concise \& Precise Justification}: The reasoning should be clear, objective, and free from unnecessary elaboration.
\end{enumerate}

\end{tcolorbox}
\caption{Prompt used by GPT-4o for NLI-based evaluation of DP.P and r-DP.R metrics.}
\label{fig:DP_P_prompt}
\end{figure*}

\begin{figure*}[t]
\begin{tcolorbox}[colback=white, colframe=black, title=Prompt for NLI Score of Dialogue Precision(r-DP.P) Based on Given Profile]

\textbf{Role} \\
You are an expert in evaluating consistency between \textbf{a given dialogue history (Source)} and a corresponding \textbf{user profile description (Target)}. Your task is to determine whether the Target aligns with, contradicts, or is ambiguous in relation to the Source.

\textbf{Task Instructions:} \\
For each \textbf{Source-Target} pair, determine the relationship using the following scoring criteria:
\begin{itemize}
    \item \textbf{Score 1}: The Target is consistent with the Source (the interpretation can be inferred from the Source).
    \item \textbf{Score -1}: The Target conflicts with the Source (the interpretation contradicts the Source).
    \item \textbf{Score 0}: The relationship is unclear or ambiguous (insufficient evidence to infer consistency or contradiction).
\end{itemize}

\textbf{Output Format:} \\
Provide your result in the following JSON format:
\begin{verbatim}
{
  "score": <score>, 
  "reason": "<concise explanation of the reasoning>"
}
\end{verbatim}

\textbf{Example :} \\
\textbf{Source}: 
\begin{itemize}
    \item [User](Turn-1): Show me how to implement a toy version of a relational database.
    \item [User](Turn-2): Thanks a lot!
\end{itemize}

\textbf{Target}: You are polite.

\textbf{Output}:
\begin{verbatim}
{
  "score": 1, 
  "reason": "The user's expression of gratitude in Turn-2 indicates politeness, 
  which aligns with the Target."
}
\end{verbatim}

\textbf{Guidelines:}
\begin{enumerate}
    \item \textbf{Contextual Inference}: Do not evaluate the Target \textbf{in isolation}. Instead, determine its logical relationship to the Source, considering both explicit statements and reasonable implications.
    \item \textbf{Concise \& Precise Justification}: The reasoning should be clear, objective, and free from unnecessary elaboration.
\end{enumerate}

\end{tcolorbox}

\caption{Prompt used by GPT-4o for NLI-based evaluation of DP.R and r-DP.P metrics.}
\label{fig:DP_R_prompt}
\end{figure*}

\begin{figure*}[t] 
\begin{tcolorbox}[colback=white, colframe=black, title=Prompt for Subjective Characteristics Score (SC.Score) in Consistency Evaluation]

\textbf{Task Description} \\
You are tasked with evaluating the quality of user responses in real human-LLM interactions. Specifically, you will assess the degree to which a given response (Target) aligns with a predefined personality profile, tone, and linguistic characteristics (Source). Additionally, you must consider the naturalness and authenticity of the Target, ensuring it reflects genuine human conversational patterns.

\textbf{Evaluation Criteria} \\
Your assessment will focus on two primary dimensions:  
\begin{enumerate}
    \item \textbf{Human-Likeness}: The extent to which Target exhibits natural human language, characterized by appropriate syntax, tone, and conversational flow.  
    \item \textbf{Alignment with Source}: The degree to which the Target adheres to the personality traits, tone, and linguistic features specified in the Source.  
\end{enumerate}

\textbf{Scoring Guidelines} \\
Assign a score from 1 to 5 based on the following criteria:  
\begin{itemize}
    \item \textbf{Score 5}: The Target fully aligns with the Source and demonstrates exceptional human-likeness.  
    \item \textbf{Score 3}: The relationship between the Target and Source is ambiguous or unclear, lacking sufficient evidence for alignment or contradiction.  
    \item \textbf{Score 1}: The Target significantly deviates from the Source or lacks human-likeness, rendering it unnatural or inconsistent.  
\end{itemize}

\textbf{Output Requirements} \\
Provide your evaluation in the following JSON format:  
\begin{verbatim}
{
  "score": <score>,
  "reason": "<concise reason>"
}
\end{verbatim}

\textbf{Key Considerations} 
\begin{enumerate}
    \item \textbf{Contextual Inference}: Analyze both explicit content and implicit nuances in the Target to determine its alignment with the Source.  
    \item \textbf{Conciseness and Precision}: Ensure that your reasoning is clear, objective, and free of superfluous elaboration.  
    \item \textbf{Human-Likeness Emphasis}: A lack of human-likeness, even if the Target aligns with the Source, will result in a lower score.  
\end{enumerate}

\end{tcolorbox}
\caption{Prompt for evaluating consistency in subjective characteristics.}
\label{fig:sc_score}
\end{figure*}

\begin{figure*}[t] 
\begin{tcolorbox}[colback=white, colframe=black, title=Prompt for Validation Score (Val.Score) in Assessing the Quality of Subjective Characteristics in Profiles]

\textbf{Role} \\
As an expert in evaluating the \textbf{consistency} between \textbf{user utterances in a dialogue (Source)} and a provided \textbf{subjective description (Target)}, your task is to assess whether the \textbf{Target} accurately reflects the characteristics described in the \textbf{Source}, including personality traits, tone, and other relevant attributes. You will then rate this consistency on a scale from 1 to 5.

\textbf{Task Instructions} \\
For each pair of \textbf{Source-Target}, apply the following scoring criteria to determine their relationship:
\begin{itemize}
    \item \textbf{Score 5}: The \textbf{Target} completely aligns with the \textbf{Source}, with no discrepancies. The profile perfectly represents the characteristics observed in the user’s utterance.
    \item \textbf{Score 3}: Ambiguity or insufficient evidence exists, making it difficult to ascertain the relationship definitively.
    \item \textbf{Score 1}: A clear discrepancy exists, with the \textbf{Target} significantly deviating from the \textbf{Source}, indicating a mismatch in the represented characteristics.
\end{itemize}

\textbf{Output Format} \\
Your assessment should adhere to the following structured JSON format:
\begin{verbatim}
{
  "score": "<numerical score>",
  "reason": "<a succinct explanation providing justification for assigned score>"
}
\end{verbatim}

\textbf{Guidelines:}
\begin{enumerate}
    \item \textbf{Contextual Inference}: Determine the target's logical relationship to the Source, considering both explicit statements and reasonable implications.
    \item \textbf{Concise \& Precise Justification}: The reasoning should be clear, objective, and free from unnecessary elaboration.
\end{enumerate}

\end{tcolorbox}
\caption{Prompt for validation score (Val.Score) in assessing the quality of subjective characteristics in profiles.}
\label{fig:val_score}
\end{figure*}

\end{CJK}
\end{document}